\documentclass[manuscript=article]{achemso}
\usepackage[utf8]{inputenc}
\usepackage[T1]{fontenc}
\usepackage{lmodern}
\usepackage{amsmath,amssymb}
\usepackage{booktabs,array,longtable,multirow}
\usepackage{graphicx}
\usepackage{float,placeins}
\usepackage{url}
\usepackage{hyperref}
\hypersetup{hidelinks}
\setkeys{acs}{articletitle=true}
\AtBeginDocument{\singlespacing}

\newcommand{\model}{PRISM}

\newcommand{\sd}[2]{\ensuremath{#1\,\pm\,#2}}
\newcommand{\sdbest}[2]{\ensuremath{\mathbf{#1}\,\pm\,\mathbf{#2}}}\newcommand{\sdsecond}[2]{\underline{\ensuremath{#1\,\pm\,#2}}}\newcommand{\papertitle}{Atom-Wise Allocation of Message-Passing Scale for Molecular Property Prediction}

\makeatletter
\renewcommand*\acs@author@list@main{\advance\@tempcnta\@ne\relax
  \def\acs@author@footnotes{}\acs@author@list@and
  \space
  \@nameuse{@author@\@roman\@tempcnta}\acs@author@star
  \acs@author@affil
  \acs@author@affil@also
  \acs@author@affil@alt
  \ifx\@empty\acs@author@footnotes\else
    \textsuperscript{\acs@author@footnotes}\fi
  \acs@author@list@comma
  \ifnum\@tempcnta<\acs@author@cnt\relax
    \expandafter\acs@author@list@main
  \fi
}
\makeatother

\author{Regina Qianru Zhang}
\affiliation{Biomedical Imaging Institute and Department of Radiology and Biomedical Imaging, Yale University, New Haven, CT 06520, USA}
\altaffiliation{These authors contributed equally.}
\author{Honggang Wen}
\affiliation{Department of Computing and Data Science, The University of Hong Kong, Pokfulam Road, Hong Kong SAR, China}
\altaffiliation{These authors contributed equally.}
\author{Zongru Li}
\affiliation{Department of Computing and Data Science, The University of Hong Kong, Pokfulam Road, Hong Kong SAR, China}
\altaffiliation{These authors contributed equally.}
\author{Jinsong Ouyang}
\affiliation{Biomedical Imaging Institute and Department of Radiology and Biomedical Imaging, Yale University, New Haven, CT 06520, USA}
\author{Zhiqin Chu}
\affiliation{Department of Electrical and Electronic Engineering, The University of Hong Kong, Pokfulam Road, Hong Kong SAR, China}
\author{Pietro Li\`o}
\affiliation{Department of Computer Science and Technology, University of Cambridge, Cambridge CB3 0FD, United Kingdom}
\email{pl219@cam.ac.uk}
\author{Xiaofeng Liu}
\affiliation{Biomedical Imaging Institute and Department of Radiology and Biomedical Imaging, Yale University, New Haven, CT 06520, USA}
\email{xiaofeng.liu@yale.edu}
\author{Siu-Ming Yiu}
\affiliation{Department of Computing and Data Science, The University of Hong Kong, Pokfulam Road, Hong Kong SAR, China}
\email{smyiu@cs.hku.hk}

\title{\papertitle}
\begin{document}
\begin{abstract}
Molecular properties depend on chemical structure at different ranges, from local polar groups governing solvation to extended substructures shaping bioactivity, yet graph neural networks usually fix one propagation scale for every atom. We present PRISM (Per-atom Routing over Intermediate States of Message Passing), in which each atom's encoded features produce a bounded scalar that positions a unimodal distribution over retained message-passing states and matching circular-fingerprint radii. Because one coordinate translates the whole profile, allocated depth responds to the routing scalar in closed form: at fixed temperature it is nondecreasing, its sensitivity scales with the profile variance, and it saturates in the clipped regions. Beyond providing closed-form analyzability, this constrained one-parameter profile matches or significantly outperforms parameter-matched, unconstrained node-wise gates and Jumping Knowledge attention. Under scaffold partitions with equal tuning budgets against eight graph, fingerprint, and geometric baselines, PRISM trained from scratch without three-dimensional coordinates achieves the lowest mean error on aqueous solubility and hydration free energy and strong accuracy on BACE and BBBP. As a general layer-combination rule, PRISM transfers directly to GCN, GIN, and GAT backbones, reducing hydration free energy error by up to 29\% relative to last-layer readout. Applied to a pre-trained GIN, it improves on last-layer readout on four of six endpoints and gives the lowest lipophilicity error overall. In trained models, allocated depth differs systematically between atom types in a task-dependent way, and permuting profiles among a molecule's atoms raises prediction error on several endpoints, so the rule acts as a learned counterpart of atom typing.
\end{abstract}

\section{Introduction}
A molecular property depends on chemical information gathered over different ranges. Aqueous solubility and lipophilicity respond to whole-molecule descriptors such as size, polarity, aromatic proportion, and flexibility\cite{delaney2004esol,wu2018moleculenet}, whereas hydration free energy is dominated by the identity and solvent exposure of a few polar groups\cite{mobley2014freesolv}. Circular fingerprints handle this by enumerating atom-centered substructures at several neighborhood radii and leaving the downstream model to decide which radii matter. Learned fingerprints and message-passing networks make the same neighborhood transformations trainable\cite{rogers2010ecfp,duvenaud2015fingerprints,gilmer2017neural}. The radius at which an atom's environment becomes informative is a modeling choice, and there is no reason to expect one choice to suit every atom.

This choice matters for molecules for three reasons. First, molecular graphs are small and sparse, and the longest shortest path in a drug-like molecule is typically below about 20 bonds. Useful propagation scales therefore differ by one or two hops, the range over which circular-fingerprint radii are chosen, and a single global depth that is one hop too large for one atom is one hop too small for another. Second, atoms in a molecule play chemically distinct roles. Classical additive models of physicochemical properties assign each atom a contribution that depends on its type and local neighborhood, as in Wildman and Crippen $\log P$ atom types and Benson group additivity\cite{wildman1999crippen,benson1958additivity}. In both, the atom's own type determines how much of its environment is relevant. Third, the endpoints studied here mix such additive group contributions with whole-molecule terms such as size, so the same molecule contains atoms whose useful information is local and atoms whose useful information is extended. An allocation conditioned on atom type serves as a learnable counterpart of atom typing, and the experiments below test this interpretation directly by disrupting the atom-to-profile correspondence and by constructing a benchmark whose target rewards atom-dependent radii.

Message-passing architectures usually make that choice once, globally. A network with $L$ layers reads its final layer, so every atom is described at the same propagation scale, and any local distinction the readout needs must survive $L$ rounds of neighbor mixing. Deeper networks do not resolve this. Repeated propagation drives atom representations toward one another\cite{li2018deeper,oono2020graph}, and narrow regions of a graph compress the information passing through them\cite{alon2021bottleneck}. Retaining intermediate layers and combining them is the standard response\cite{xu2018jumping,chen2020gcnii}, but a uniform average, or a single learned weight profile shared by all atoms, still applies one combination throughout the molecule. It cannot weight a heteroatom toward its immediate environment while weighting an aromatic carbon toward a more extended one.

Several established methods address parts of this. Jumping Knowledge networks aggregate across neighborhood ranges with node-dependent, essentially unconstrained selection over layers, and adaptive propagation gives nodes learned halting decisions that stop computation early\cite{xu2018jumping,spinelli2020adaptive}. Residual and identity mappings make deep stacks trainable, and curvature-based rewiring modifies the graph itself to relieve bottlenecks\cite{chen2020gcnii,topping2022curvature}. Multi-scale and multimodal molecular encoders pursue a related goal from a different direction. MMGIN combines a graph isomorphism network over the molecular graph with a parallel fingerprint channel, so that substructure information at fixed fingerprint radii complements learned graph features in multitask toxicity prediction\cite{wang2024mmgin}. MSIGR-PLA integrates multi-scale protein-ligand interaction features from graph convolution, cross-attention, and graph transformer modules with global sequence- and structure-level representations for affinity prediction\cite{zhang2026msigrpla}. Both enrich the molecular representation by fusing several fixed views or scales, with the fusion applied identically to every atom. PRISM addresses a narrower and complementary question by asking how much each retained depth in a single message-passing stack should contribute to a given atom's representation. The approaches are compatible, and a fingerprint channel or an interaction module could be paired with per-atom depth allocation. Together, these established methods leave open an intermediate design point in which a per-atom allocation is more expressive than a shared profile and more constrained than free per-layer weights, with behavior that can be characterized in closed form.

\begin{figure}[!htbp]
\centering
\includegraphics[width=\textwidth]{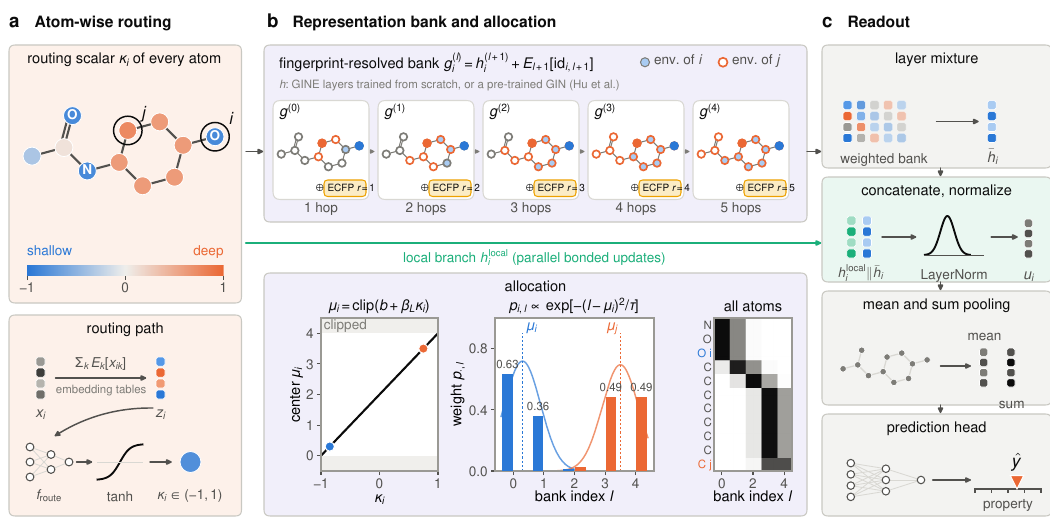}
\caption{Atom-wise allocation of propagation scale in PRISM. (a) Routing: each atom's descriptors are encoded by categorical embedding tables into $z_i$, and a routing network maps $z_i$ to a bounded scalar $\kappa_i$. Atoms of acetaminophen are colored by an illustrative $\kappa_i$ from shallow (blue) to deep (orange), and atoms $i$ (hydroxyl oxygen) and $j$ (ring carbon) are followed through the figure. (b) Fingerprint-resolved bank and allocation: the bank holds $L$ states $g_i^{(l)}$, each the representation after $l+1$ message-passing updates plus an embedding of the atom's circular-fingerprint (ECFP) environment at radius $l+1$. Filled and ringed atoms mark the radius-$(l+1)$ environments of $i$ and $j$. The updates come from GINE layers trained from scratch or from a pre-trained GIN\cite{hu2020strategies}. The lower panels show the clipped map from $\kappa_i$ to the center $\mu_i$, the Gaussian layer weights $p_{i,l}$ of $i$ and $j$, and the weights of every atom of the molecule. (c) Readout: the weighted mixture $\bar h_i$ is concatenated with the local-branch representation, normalized, pooled by mean and sum over atoms, and passed to the prediction head. The $\kappa_i$ values and profiles are illustrative.}
\label{fig:architecture}
\end{figure}

PRISM (Per-atom Routing over Intermediate States of Message Passing) occupies that point. Just as an optical prism separates light into its components, the method separates a molecule's message-passing computation into retained propagation scales and recombines them atom by atom. The core contribution is a layer-combination rule, scalar-parameterized routing over a representation bank, that sits on top of any message-passing operator. It is not a new operator, and it transfers to GCN, GIN, and GAT backbones (see Results). As outlined in Figure~\ref{fig:architecture}, each atom's encoded input features produce one bounded scalar, which positions a unimodal distribution over a bank of retained post-update states. Because a single coordinate moves the whole profile, neighboring layer weights stay coupled, and the scalar has a derivable, monotone effect on the atom's allocated depth, including in the boundary regions where clipping saturates the response. By default, the scalar is computed from the atom's input encoding rather than from an aggregated neighborhood, so the learned allocation is conditioned on atom type and local descriptors rather than on the surrounding environment, and the unimodal family excludes arbitrary layer profiles. Three further components accompany the allocation in the full architecture: a fingerprint-resolved bank that pairs each message-passing depth with the corresponding circular-fingerprint radius, a parallel local branch regularized by an anti-collapse feature-energy hinge (with a bond-smoothness penalty on the routing scalar evaluated in ablation), and a readout concatenating mean and sum pooling, whose complementary roles in invariant graph prediction are established\cite{zaheer2017deepsets,xu2018powerful}. These address distinct constraints, so any predictive gain is a property of the combination, and the ablations below separate their contributions.

This work asks three questions. Does atom-dependent allocation of propagation scale improve molecular property prediction in a matched comparison, and does it improve on the last-layer readout of a pre-trained network? Which components produce that improvement? And what does the design cost in computation? We answer the first with physicochemical regression on ESOL, FreeSolv, and Lipophilicity and classification on BACE, BBBP, and SIDER under scaffold partitions, against graph, geometric, message-passing, and fingerprint baselines with reported parameter counts, matched tuning budgets, and paired tests over the partitions, and by applying PRISM allocation to the pre-trained GIN of Hu et al.\cite{hu2020strategies} in place of its last-layer readout. We answer the second with component removals, node-dependent layer-weighting alternatives including gates and Jumping Knowledge, a transfer of the allocation module to three standard backbones, a within-molecule shuffling of the learned profiles, and a synthetic benchmark that rewards atom-dependent radii. We answer the third with parameter, FLOP, time, and memory measurements and with a depth sweep on Peptides-struct\cite{dwivedi2022lrgb}.

\section{Methods}
\subsection{Representation Bank}
PRISM separates computing representations from choosing how to combine them. A topological branch produces states at successive propagation depths and retains all of them, and an allocation module decides, per atom, how much each state contributes. Early states reach the readout directly, even when deeper states are also computed.

Let $\mathcal G=(\mathcal V,\mathcal E)$ be a molecular graph with atom features $x_i$ and bond features $e_{ij}$. Message passing supplies permutation-equivariant node updates of the usual form\cite{gilmer2017neural}
\begin{equation}
 h_i^{(l+1)}=\Phi_l\left(h_i^{(l)},\operatorname*{AGG}_{j\in\mathcal N(i)}\Psi_l(h_i^{(l)},h_j^{(l)},e_{ij})\right).
\label{eq:mp}
\end{equation}
PRISM has a local branch and a topological branch. The topological branch initializes from the atom encoding plus a learned degree embedding and applies residual graph updates. In all molecular experiments, each update is a pre-normalized residual GINE convolution with bond features\cite{hu2020strategies}, $h_i^{(l+1)}=h_i^{(l)}+\alpha\,\mathrm{GINE}\bigl(\mathrm{LayerNorm}(h^{(l)})\bigr)_i$. Its update network is a two-layer GELU perceptron of width $2H$ with a learned residual scale $\alpha$. Its bank is $g_i^{(l)}=h_i^{(l+1)}$ for $l=0,\ldots,L-1$ (augmented with a radius-matched substructure embedding in the fingerprint-resolved bank of eq~\ref{eq:fpbank}), so every stored state follows a topological update and the unpropagated input is not a bank entry. Consequently $L$ counts retained post-update states, and a mean index of zero corresponds to the first topological transition rather than to no propagation. The local branch applies a stack of the same pre-normalized residual GINE updates of tuned depth (reported with the bank depth $L$ in Supporting Information Table~S1) to supply a parallel representation $h_i^{\mathrm{local}}$ retained alongside the allocated mixture, and the two branches are not constrained to encode disjoint information.

Because the bank only stores post-update states, the allocation module is independent of the operator that produces them. Replacing the update in eq~\ref{eq:mp} with a GCN, GIN, or GAT layer\cite{kipf2017semi,xu2018powerful,velivckovic2017graph} leaves the routing network, the weights in eq~\ref{eq:weights}, and the mixture in eq~\ref{eq:aggregate} unchanged, allowing us to test whether the allocation transfers across backbones (see Results). The same independence allows the bank to be a pre-trained network. We use the five-layer GIN of Hu et al.\cite{hu2020strategies}, pre-trained by context prediction on 2 million ZINC15 molecules followed by supervised multitask pre-training on about 456,000 ChEMBL molecules, with its original atom (element, chirality) and bond (type, direction) inputs and its released weights. Its five post-update layer states form the bank, the routing network reads its input embedding, and the whole network is fine-tuned end to end. Four readouts of this bank are compared with the same 10 trial settings per dataset (dropout $\{0,0.1,0.2,0.3,0.5\}$, learning rate $\{10^{-4},3\times10^{-4},10^{-3}\}$, and batch size $\{32,64\}$, with width 300 and depth 5 fixed by the checkpoint): the last layer, which is the standard fine-tuning readout, node-wise Jumping Knowledge attention, PRISM allocation, and PRISM allocation over the fingerprint-resolved bank of eq~\ref{eq:fpbank}. Which pre-trained PRISM variant (with or without the fingerprint-resolved bank) enters Tables~\ref{tab:scaffold} and \ref{tab:classification} was decided before testing, based on the mean validation rank across the six datasets.

\subsection{Scalar-Parameterized Allocation}
The encoded atom features determine a bounded scalar before any neighbor aggregation,
\begin{align}
 z_i&=f_{\mathrm{atom}}(x_i),\qquad
 \kappa_i=\tanh\!\left(f_{\mathrm{route}}(z_i)\right),\nonumber\\
 c_i&=b+\beta\kappa_i,\qquad
 \mu_i=\operatorname{clip}(c_i,0,L-1).
\label{eq:center}
\end{align}
Here $b$ is a shared reference center, $\beta>0$ is the scale parameter, and $\beta\kappa_i$ is the atom-dependent shift. Because $\kappa_i\in(-1,1)$, the unclipped center lies in $(b-\beta,b+\beta)$, an interval of width $2\beta$ index units. With a fixed scale $\beta=2$, this interval has width 4, which covers the whole bank for $L\le5$ and only part of the bank for deeper stacks. With $L=15$, for example, centers under $\beta=2$ can reach at most 4 of the 14 index units around $b$, and the remaining layers receive weight only through the tails of the profile. We report $b$ for every configuration (Supporting Information Table~S1), use the depth-scaled coefficient $\beta_L=(L-1)/2$ with $b=(L-1)/2$ (which equals $\beta=2$ when $L=5$) for the molecular benchmarks, and compare fixed $\beta=2$ with $\beta_L=(L-1)/2$ across the Peptides-struct depth sweep. Because $\beta>0$, a larger routing scalar moves the center toward later states. Clipping confines the center to the available bank and does not halt computation. With learned $t$ and effective temperature $\tau=\max(|t|,0.1)$, the normalized profile is
\begin{equation}
 p_{i,l}=\frac{\exp[-(l-\mu_i)^2/\tau]}{\sum_{k=0}^{L-1}\exp[-(k-\mu_i)^2/\tau]}.
\label{eq:weights}
\end{equation}
The squared distance concentrates weight near the center $\mu_i$ and decays smoothly away from it. Small $\tau$ pins the profile to one or two layers, and large $\tau$ spreads it across the bank. At fixed $b$ and $\tau$, one scalar translates the entire profile instead of setting each layer weight independently. Normalization makes the mixture a weighted average, so its magnitude does not grow simply because more states are included.

The mixture and the allocated index are
\begin{equation}
 \bar h_i=\sum_{l=0}^{L-1}p_{i,l}g_i^{(l)},\qquad
 d_i=\sum_{l=0}^{L-1}l\,p_{i,l}.
\label{eq:aggregate}
\end{equation}
We report $d_i$ as a zero-based index, so $d_i+1$ is the weighted count of topological transitions. The routing scalar $\kappa_i$ and allocated index $d_i$ are learned routing variables, not measured receptive fields. Because the bank is discrete and finite, the expected index can differ from the center, most visibly near the upper boundary.

Away from the clipping boundaries, and holding the bank, $b$, and $\tau$ fixed,
\begin{equation}
 \frac{\partial d_i}{\partial\kappa_i}=
 \begin{cases}
 \displaystyle\frac{2\beta}{\tau}\operatorname{Var}_{p_i}(l),&0<c_i<L-1,\\[4pt]
 0,&c_i<0\ \text{or}\ c_i>L-1,
 \end{cases}
\label{eq:monotone}
\end{equation}
giving an interior coefficient of $4/\tau$ when $\beta=2$. The derivative is generally undefined at a boundary, and one-sided limits are given in Supporting Information Section S1. Allocated depth is globally nondecreasing in $\kappa_i$, strictly increasing in the unclipped region for $L>1$ and finite temperature, and constant once the center is clipped. A profile already spread over several indices can shift its expectation readily, whereas a concentrated profile has little weight to redistribute. Equation~\ref{eq:monotone} characterizes the allocation, not its effect on accuracy.

Two qualifications apply to this characterization. First, $\kappa_i$ is a deterministic function of $x_i$, so identical atom-feature vectors receive identical profiles within a fitted model. The allocation is conditioned on the atom's own descriptors, and an association between $\kappa_i$ and, say, aromaticity reflects the input encoding rather than the surrounding environment. Second, monotonicity of $d_i$ does not transfer through the nonlinear prediction head. Relative to Jumping Knowledge aggregation and adaptive propagation, this construction supplies a restricted distribution over a fully evaluated bank and implements no layer-wise stopping decision\cite{xu2018jumping,spinelli2020adaptive}.

Conditioning on the atom's own descriptors is intentional. It makes the allocation a learnable counterpart of atom typing (Introduction) and keeps $\kappa_i$, and hence the closed-form depth response, independent of the bank. To test whether environment information would help, we also evaluate an environment-aware variant, PRISM-env, in which the routing network reads the first post-update state, $\kappa_i=\tanh(f_{\mathrm{route}}(h_i^{(1)}))$, so that atoms with identical descriptors in different one-hop environments can receive different profiles. In PRISM-env, eqs~\ref{eq:weights} to \ref{eq:monotone} still hold conditionally on $h^{(1)}$, but $\kappa_i$ is no longer a function of $x_i$ alone.

\subsection{Featurization, Allocation Defaults, and Fingerprint-Resolved Bank}
\label{sec:featurization}
Atoms and bonds carry the nine and three integer descriptors of the PyTorch Geometric MoleculeNet featurization. Each descriptor column is encoded by a separate learned embedding table and the embeddings are summed, rather than applying a linear map to the raw integers. For every graph baseline both encoders are run and the one with the better validation score is reported (Supporting Information), so that no baseline is disadvantaged by the featurization. In all molecular experiments the allocation uses the depth-scaled coefficient $\beta_L=(L-1)/2$ with $b=(L-1)/2$ and an initial temperature $t=0.5$. These defaults were selected, on validation data only, among 12 combinations of $\beta\in\{2,\beta_L\}$, initial $t\in\{0.5,1,2\}$, and an optional normalization of the routing input.

Circular fingerprints describe each atom by the substructure centered on it at radius $r=0,1,2,\ldots$, and the Introduction motivates PRISM as a learned choice among such radii. The fingerprint-resolved bank makes this correspondence explicit. Bank state $l$, which follows $l+1$ topological updates, is augmented with an embedding of the atom's Morgan environment identifier at radius $l+1$\cite{rogers2010ecfp},
\begin{equation}
 g_i^{(l)}=h_i^{(l+1)}+E_{l+1}\bigl[\mathrm{id}_{i,\,l+1}\bigr],
\label{eq:fpbank}
\end{equation}
where $\mathrm{id}_{i,r}$ is the RDKit identifier of the radius-$r$ environment of atom $i$ (repeated from the largest smaller radius when RDKit merges an environment), hashed into 4096 buckets per radius. The allocation of eqs~\ref{eq:center} to \ref{eq:aggregate} is unchanged, so $p_{i,l}$ now also selects which substructure radius the readout sees for atom $i$.

\subsection{Graph Regularization and Molecular Readout}
Graph smoothness penalties are a standard way to discourage variation across connected samples\cite{belkin2006manifold}. The scalar-field term is
\begin{equation}
 \mathcal L_{\kappa}=\frac{a}{|\mathcal E_d|}
 \sum_{(i,j)\in\mathcal E_d}(\kappa_i-\kappa_j)^2,
\label{eq:lkappa}
\end{equation}
with $\mathcal E_d$ the directed edge list and a default strength $a=0.2$. This discrete Dirichlet penalty makes bonded atoms prefer similar routing values unless the task objective outweighs the difference. It expresses a soft preference for coherence along bonds without forcing a shared profile.

For a hidden representation $H$, define the feature-energy statistic
\begin{equation}
 E(H)=\frac{1}{|\mathcal E_d|}\sum_{(i,j)\in\mathcal E_d}\|h_i-h_j\|_2^2 ,
\label{eq:energy}
\end{equation}
evaluated in the local branch before and after its updates. The implemented feature term is an anti-collapse hinge,
\begin{equation}
 \mathcal L_{\mathrm{feat}}=\max\{0,\,0.8E(H_{\mathrm{in}})-E(H_{\mathrm{local}})\},
\label{eq:lfeat}
\end{equation}
positive when output feature energy falls below 80\% of its input value and zero at or above that threshold. Feature Dirichlet energy is the standard quantity for measuring the collapse of node representations under repeated propagation\cite{rusch2023survey}, and the hinge discourages that collapse in the local branch without prescribing any particular value. Smoothing a scalar field and preserving representation differences serve different purposes, which is why the two penalties act on different objects. The total objective is
\begin{equation}
 \mathcal L=\mathcal L_{\mathrm{task}}+\lambda\left(\mathcal L_{\kappa}+0.1\,\mathcal L_{\mathrm{feat}}\right),
\label{eq:total}
\end{equation}
with $\mathcal L_{\mathrm{task}}$ the task loss (mean squared error for regression, binary cross-entropy with masking for multitask classification, and mean absolute error for Peptides-struct) and an outer coefficient $\lambda$ selected per dataset.

The scalar-field penalty $\mathcal L_\kappa$ is omitted from the final molecular models ($\mathcal L=\mathcal L_{\mathrm{task}}+0.1\,\lambda\mathcal L_{\mathrm{feat}}$) because, at the strengths selected on validation, it compressed the routing scalar to a nearly constant value within each model (SD of $c_i$ between 0.001 and 0.04 index units on ESOL, BACE, and BBBP). Table~\ref{tab:ablation} and Supporting Information Table~S10 report the effect of including $\mathcal L_\kappa$.

The forward pass normalizes the concatenation of local and topological representations and pools it twice,
\begin{equation}
 u_i=\operatorname{LayerNorm}\!\left(h_i^{\mathrm{local}}\Vert\bar h_i\right),\qquad
 r=\left(\frac{1}{|\mathcal V|}\sum_i u_i\right)\Vert\left(\sum_i u_i\right),
\label{eq:readout}
\end{equation}
supplying $r$ to a feed-forward head. Mean pooling summarizes average atom features, and sum pooling additionally retains accumulated magnitude. For a fixed multiset of atom vectors, duplicating every vector leaves the mean unchanged but doubles the sum, which is why mean pooling alone can hide multiplicity. Both summaries are permutation-invariant\cite{zaheer2017deepsets,xu2018powerful}, and their concatenation lets the head use whichever the task requires.

\subsection{Layer-Weighting Alternatives and Allocation Controls}
\label{sec:controls}
All alternatives below modify only how the bank states $g_i^{(l)}$ are combined into $\bar h_i$ in eq~\ref{eq:aggregate}. The input encoding, degree embedding, bank, local branch, regularizers (where applicable), mean-and-sum readout, prediction head, partitions, and training schedule are those of the full model. Two schemes replace $p_{i,l}$ with unconstrained node-dependent layer weights:
(i) a node-wise layer gate, $p_{i,l}=\operatorname{softmax}_l\!\left(\mathrm{MLP}(z_i)\right)_l$, which uses the same input as PRISM's routing network but outputs $L$ free logits instead of one scalar, and
(ii) node-wise Jumping Knowledge attention, $p_{i,l}=\operatorname{softmax}_l\!\left(a^\top g_i^{(l)}\right)$, whose scores depend on the bank states themselves\cite{xu2018jumping}.
Two further Jumping Knowledge schemes replace the weighted sum in eq~\ref{eq:aggregate} directly: (iii) layer concatenation followed by a linear projection to the hidden size and (iv) element-wise max pooling over layers\cite{xu2018jumping}.
The hidden width of the routing or scoring network in (i) and (ii) is chosen so that the total parameter count matches full PRISM to within 0.3\% (Supporting Information Section~S3 and Table~S10). Together with PRISM-env, these alternatives separate three properties of the allocation: whether it is atom-dependent, whether it is constrained to a unimodal one-parameter family, and what information it is conditioned on.

Two further controls test whether the particular assignment of profiles to atoms matters in a trained model. With the model, the bank, and the routing outputs fixed, we evaluate the test set (a) after permuting the layer-weight vectors $p_i$ uniformly at random among the atoms of each molecule, which preserves the molecule's multiset of profiles and its mean allocated depth while breaking the atom-to-profile correspondence, and (b) after replacing every $p_i$ in a molecule by the molecular average $\bar p=|\mathcal V|^{-1}\sum_i p_i$, which removes within-molecule heterogeneity entirely. Each permutation is repeated with 20 random seeds per trained model, and the change in test error relative to the unperturbed model is reported over the five trained models. Molecules with a single heavy atom are left unchanged.

Finally, we define a synthetic atom-dependent-radius (ADR) target built to reward atom-specific propagation scales on the molecules of ESOL and Lipophilicity, pooled and deduplicated by canonical SMILES (5267 molecules). For each heavy atom $i$ let $n_i^{(r)}$ be the number of heavy atoms at shortest-path distance $r$ from $i$. The target is
\begin{equation}
y_{\mathrm{ADR}}=\sum_{i\in\mathrm{hetero}}n_i^{(1)}+\sum_{i\in\mathrm{carbon}}n_i^{(r_C)},
\label{eq:adr}
\end{equation}
so heteroatoms contribute information at one hop and carbons at $r_C$ hops, with $r_C\in\{2,3,4\}$. The target is computed from the graph, so the only source of error is the model. A model with one global propagation scale must represent both radii in a single state, whereas atom-type-conditioned allocation can read each atom at its own radius.

\subsection{Computational Cost}
At fixed hidden dimension, building and mixing the bank costs $\mathcal O(L(|\mathcal E|+|\mathcal V|))$, with the mixture adding $\mathcal O(L|\mathcal V|)$ beyond the backbone. There is no dense all-pairs attention matrix. Allocating weight to earlier states does not avoid computing later ones, because every bank entry is evaluated regardless of the profile, so the mechanism is not an early exit and yields no computational saving. Measured time and memory are reported alongside accuracy together with the depth sweep in Results, and parameter counts, floating-point operations (FLOPs) per molecule, and training time per epoch are reported for every model in Table~\ref{tab:cost}. FLOPs are counted with \texttt{torch.utils.flop\_counter} (PyTorch 2.5) on a forward pass over the ESOL test set and divided by the number of molecules.

\subsection{Datasets and Evaluation Protocol}
\label{sec:protocol}
The primary regression comparison uses the MoleculeNet versions of ESOL (1128 molecules), FreeSolv (642), and Lipophilicity (4200), with RMSE in $\log(\mathrm{mol/L})$, standardized units, and $\log D$ respectively. FreeSolv hydration free energies are standardized by their standard deviation over the dataset, 3.84\,kcal/mol, so one RMSE unit corresponds to 3.84\,kcal/mol\cite{wu2018moleculenet,delaney2004esol,mobley2014freesolv}. Each regression dataset is evaluated under two partitions at 80/10/10 proportions. In the random partition, molecules are assigned at random, redrawn for each of five seeds. In the scaffold partition, molecules are grouped by Bemis and Murcko scaffold\cite{bemis1996scaffold}, and whole scaffold groups are assigned to training, validation, and test sets, with the group order randomized by seed so that five distinct scaffold partitions are obtained\cite{yang2019analyzing}. For both protocols the five partitions are generated once, stored as index files, and reused unchanged by every model, so all models are trained and tested on identical molecules. Scaffold-partition errors are expected to be higher than random-partition errors because test scaffolds are absent from training, and benchmark studies show that the splitting strategy can change measured performance substantially\cite{li2026survey}.

The comparison includes graph baselines (GCN, GAT, GIN\cite{kipf2017semi,velivckovic2017graph,xu2018powerful}, and KA-GCN\cite{li2025kolmogorov}), a directed message-passing neural network (D-MPNN, as implemented in Chemprop\cite{yang2019analyzing}), geometric baselines (EquiformerV2 and MACE\cite{liao2023equiformerv2,batatia2022mace}), and a fingerprint baseline, a random forest\cite{breiman2001rf} on 2048-bit Morgan count fingerprints of radius 2\cite{rogers2010ecfp}. The geometric models receive three-dimensional coordinates. For each molecule, 10 conformers are generated with RDKit ETKDGv3\cite{riniker2015etkdg} from the MoleculeNet SMILES with explicit hydrogens and random seed 0, each is optimized with MMFF94\cite{halgren1996mmff} (UFF where MMFF94 parameters are unavailable), and the lowest-energy conformer is retained. One conformer per molecule is computed once and shared by all seeds and partitions. ETKDG embedding succeeded for every ESOL, FreeSolv, and BACE molecule. For 2 Lipophilicity, 4 BBBP, and 31 SIDER molecules it succeeded only from random initial coordinates. For 1 BBBP and 15 SIDER molecules no three-dimensional embedding was obtained within 10\,min, and RDKit two-dimensional coordinates were used instead. UFF replaced MMFF94 for 7, 38, and 81 molecules of Lipophilicity, BBBP, and SIDER, respectively. All molecules are retained for every model. Explicit hydrogens are used during embedding and optimization and then removed so that the geometric inputs contain heavy atoms only, matching the molecular graphs. The graph-based models, including PRISM, use no coordinates.

The from-scratch graph models share the MoleculeNet atom and bond descriptors, while the pre-trained GIN, D-MPNN, and geometric baselines use their standard inputs described above. All neural models share the same stored partitions, AdamW optimizer, learning-rate schedule, early-stopping rule on validation error, and evaluation code. Hyperparameters are selected with the same budget for every model: 10 random-search trials per dataset, evaluated on the validation split of the first partition, over hidden size $\{64,128,256,300\}$, number of message-passing layers $\{2,3,4,5,6\}$, dropout $\{0,0.1,0.2,0.3\}$, learning rate $\{10^{-4},3\times10^{-4},10^{-3}\}$, and batch size $\{32,64\}$. PRISM additionally searches $\lambda\in\{0.01,0.05,0.1\}$ and three binary options: routing on the one-hop state $h_i^{(1)}$ instead of $z_i$ (PRISM-env, selected on BBBP and SIDER), a per-atom profile width $\tau_i=\max(|t_i|,0.1)$ predicted from the routing input instead of a global $\tau$ (selected on FreeSolv, BBBP, and SIDER), and concatenation of a projected 2048-bit Morgan count fingerprint into the molecular readout $r$ (selected on BACE), with trial 0 fixed to the default per-dataset configuration (on Lipophilicity, where validation selected it, hidden size 192, learning rate $2\times10^{-4}$, batch size 128, $\lambda=0.001$, and 400 epochs with patience 100, Supporting Information Table~S1). The geometric models additionally search the radial cutoff $\{4,5,6\}$\,\AA. The random forest searches the number of trees $\{200,500,1000\}$, maximum features $\{\sqrt{\cdot},0.3,1.0\}$, and minimum samples per leaf $\{1,2,4\}$ with the same number of trials, and D-MPNN uses Chemprop's hyperparameter optimization with the same trial count. The selected configuration is then trained with five seeds. Parameter counts and the selected settings are reported in Table~\ref{tab:cost} and Supporting Information Table~S1. Because tuning gives each baseline its own best setting, parameter counts are not equal. We also report a control in which GIN is widened to approximately the parameter count of PRISM, and PRISM is narrowed to approximately that of GIN.

Classification uses BACE, BBBP, and SIDER with Bemis and Murcko scaffold partitioning at the same proportions, both for the baseline comparison and for the ablations. The component ablations are reported on SIDER in the main text (Table~\ref{tab:ablation}) and, together with the node-dependent layer-weighting alternatives, on all six endpoints (with ESOL, FreeSolv, and Lipophilicity under the random partition) in Supporting Information Table~S10. Classification performance is the ROC-AUC\cite{fawcett2006roc} of the predicted probability of the positive class, $\hat p=\sigma(\text{logit})$, averaged over tasks. The output-to-label mapping is fixed by the training loss and is identical for validation and test, with checkpoints selected on validation ROC-AUC. A stratified random partition is substituted for any seed in which the scaffold-grouped validation or test fold would contain a single class (Supporting Information Section~S2).
Peptides-struct from the Long Range Graph Benchmark supplies a depth sweep at 5, 7, 9, 11, and 15 layers, reporting MAE together with time per epoch and peak GPU memory from a single run per configuration\cite{dwivedi2022lrgb}. In this sweep the allocation parameters use a learning rate of $10^{-5}$ against $10^{-3}$ for the remaining parameters and are held fixed for the first 10 of 100 epochs. The sweep is run with both $\beta=2$ and the depth-scaled $\beta_L$, with $b$ reported for each configuration. Uncertainties are standard deviations (SD) over five runs. Relative error reduction is $100(E_{\mathrm{baseline}}-E_{\mathrm{\model}})/E_{\mathrm{baseline}}$.

\section{Results}
\subsection{Behavior of the Allocation Rule and Learned Allocations}
\begin{figure}[!htbp]
\centering
\includegraphics[width=\textwidth]{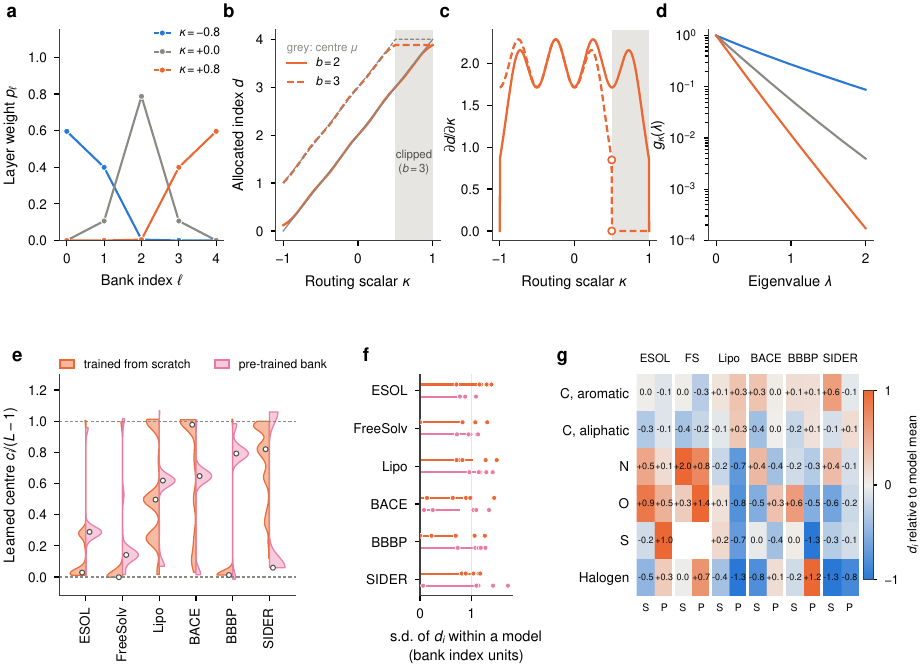}
\caption{The depth-allocation rule in closed form (a to d) and as learned by the trained models (e to g). (a to c) Evaluations of eqs~\ref{eq:center} to \ref{eq:monotone} at the five-layer initialization ($L=5$, $b=\beta_L=(L-1)/2=2$, $\tau=0.5$, solid) used by five of the six molecular models, and with the reference center moved to $b=3$ (dashed). (a) Layer-weight profiles at three routing scalars. (b) Allocated index $d$ (orange) and clipped center $\mu$ (gray) against $\kappa$, where the shaded band is the clipped region $\kappa>0.5$ for $b=3$. (c) Sensitivity $\partial d/\partial\kappa$, which vanishes on the plateau, with open circles marking the one-sided limits at the boundary. (d) Response of the same weight family under a linear heat kernel with diffusion times $t_l=l+1$, $g_\kappa(\lambda)=\sum_l p_l e^{-(l+1)\lambda}$, an analogy for a linearized operator. (e) Learned centers $c_i=b+\beta_L\kappa_i$, scaled by $L-1$, over the test atoms of one trained model per dataset (first scaffold partition), showing PRISM trained from scratch on the left (Tables~\ref{tab:scaffold} and \ref{tab:classification}) and PRISM allocation on the pre-trained GIN on the right (Table~\ref{tab:pretrained}), with dots marking medians. (f) Standard deviation (SD) of the allocated index $d_i$ over the test atoms of each trained model (dots show the five partitions and bars show the mean), from scratch (orange) and on the pre-trained bank (pink). (g) Mean allocated index of each atom type relative to the mean of its model, averaged over the five models, from scratch (S) and pre-trained (P), where FS denotes FreeSolv and blank cells indicate fewer than 50 test atoms. Numerical values are in Supporting Information Table~S5.}
\label{fig:allocation}
\end{figure}

Figure~\ref{fig:allocation} evaluates eqs~\ref{eq:center} to \ref{eq:monotone} at the five-layer initialization $L=5$, $b=\beta_L=2$, and $\tau=0.5$ (selected bank depths for all datasets are listed in Supporting Information Table~S1). Raising $\kappa$ from $-0.8$ to $0$ and $+0.8$ moves the profile toward later post-update states (Figure~\ref{fig:allocation}a), giving allocated indices of 0.409, 2.000, and 3.591. Because the distribution is normalized over a finite bank, its expected index need not equal its center, as clipping the center at four does not place all probability on layer four (Figure~\ref{fig:allocation}b). The response to $\kappa$ is positive only in the unclipped region, scaling with the variance of the depth distribution and falling to zero on the plateau (Figure~\ref{fig:allocation}c). With $b=3$, at $\kappa=0.5$ the left limit is 0.848 and the right limit is zero. At the initialization $b=\beta_L$ the unclipped range of $\mu$ is $[0,L-1]$, so a plateau is reached only as $|\kappa|\to1$. Distinct scalar values inside the plateau produce identical profiles, so raw differences in $\kappa$ do not by themselves indicate a change in allocated computation. The variance form of the derivative is the exponential-family identity for a location parameter. For the trained molecular models, the distributions of learned $\kappa_i$ and allocated indices $d_i$, the fitted $b$ and $\tau$, and the fraction of atoms whose center lies on a clipping plateau are reported per dataset and per bank depth in Supporting Information Section~S5, and the learned allocations are shown in Figure~\ref{fig:allocation}e to g. In every final model the allocation varies across atoms. Within a trained model the SD of the unclipped centers $c_i$ is 0.70 to 1.18 index units across the six datasets (with the SD of the allocated index $d_i$ in Figure~\ref{fig:allocation}f ranging from 0.65 to 1.15 units), and 83\% to 94\% of the center variation lies within molecules. The variation follows atom type in a task-dependent way. Relative to the mean of its model, nitrogen atoms are allocated 2.05 index units deeper on FreeSolv, oxygen and nitrogen atoms are allocated 0.87 and 0.53 units deeper and halogens 0.48 units shallower on ESOL, whereas on SIDER halogens are allocated 1.26 units shallower and aromatic carbons 0.60 units deeper. Between 0.9\% (ESOL) and 74.2\% (BBBP) of test atoms lie on a clipping plateau across the five scaffold partitions (Supporting Information Table~S5). PRISM allocation over the pre-trained GIN has a center SD of 0.83 to 1.35 units (and a $d_i$ SD of 0.76 to 1.20 units), with 79\% to 98\% of the center variation lying within molecules.

Under a linear heat kernel, profiles centered deeper attenuate nonzero eigenvalues more strongly (Figure~\ref{fig:allocation}d). This ordering follows from the Gaussian family and the fixed operator, and it does not show that the trained nonlinear network avoids over-smoothing.

\subsection{Comparison with Eight Baselines under Scaffold Partitions}
\begin{figure}[!htbp]
\centering
\includegraphics[width=\textwidth]{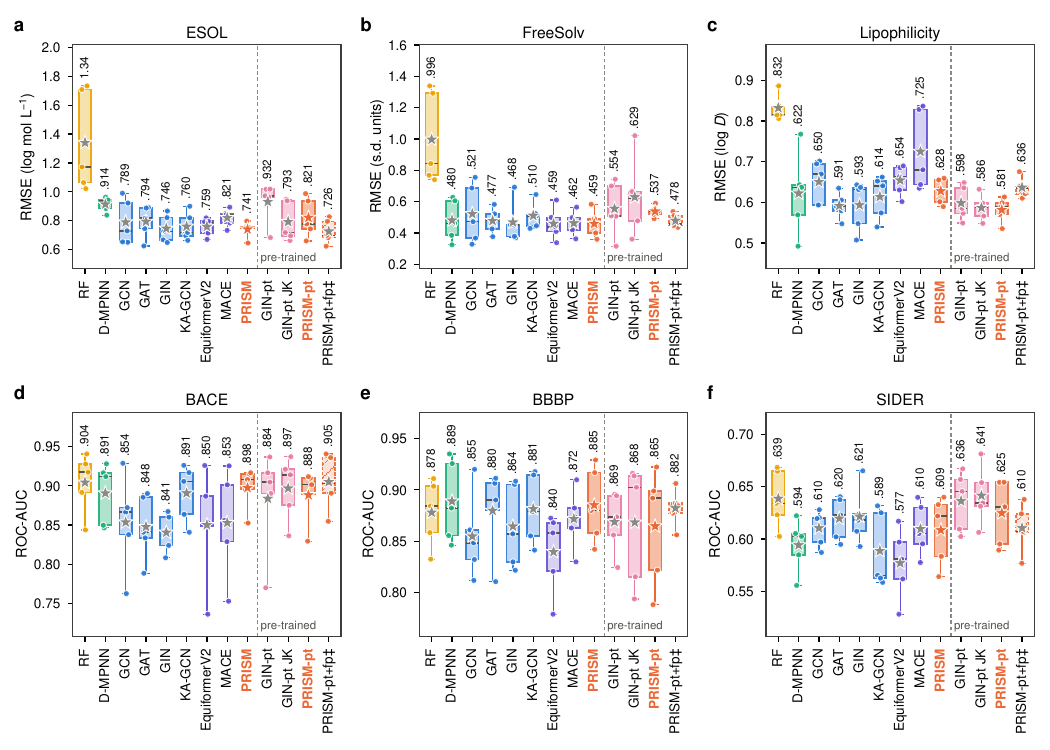}
\caption{Test performance of every model on the five scaffold partitions: (a) ESOL, (b) FreeSolv, and (c) Lipophilicity (RMSE, lower is better), and (d) BACE, (e) BBBP, and (f) SIDER (ROC-AUC, higher is better). Each box spans the five partitions, dots are the individual partitions, and the star and number above it mark the mean. PRISM is shown in orange, and the models to the right of the dashed line apply different readouts to the pre-trained GIN of Hu et al.\cite{hu2020strategies} (Table~\ref{tab:pretrained}). The hatched box shows the unselected PRISM variant with the fingerprint-resolved bank on the pre-trained stack. Models are those of Tables~\ref{tab:scaffold} and \ref{tab:classification}. FreeSolv RMSE is in units of the dataset SD (1 unit = 3.84\,kcal/mol).}
\label{fig:regression}
\end{figure}

We compare PRISM against eight graph, message-passing, fingerprint, and geometric baselines on three physicochemical and three classification endpoints, using five stored Bemis and Murcko scaffold partitions that are identical for every model and the same 10-trial tuning budget (Methods, Tables~\ref{tab:scaffold} and \ref{tab:classification}, Figure~\ref{fig:regression}). On the regression endpoints, PRISM has the lowest mean RMSE on ESOL, $0.741\pm0.057$ vs $0.746\pm0.095$ for GIN, and on FreeSolv, $0.459\pm0.086$ vs $0.459\pm0.103$ for EquiformerV2, a margin of $2\times10^{-4}$ that amounts to a tie. On Lipophilicity it is fifth, with RMSE $0.628\pm0.031$ vs $0.591\pm0.032$ for GAT, 6.3\% above the best model trained from scratch, while PRISM allocation over the pre-trained GIN reaches $0.581\pm0.030$, the lowest Lipophilicity error of all models (Table~\ref{tab:pretrained}). Paired tests over the five partitions (Supporting Information Table~S8) find no baseline significantly better than PRISM on ESOL or FreeSolv, and only GAT significantly better on Lipophilicity. The Morgan-fingerprint random forest has the highest error on all three regression endpoints.

\begin{table}[tb]
\centering
\caption{Physicochemical regression under scaffold partitions.}
\label{tab:scaffold}
{\small\setlength{\tabcolsep}{4pt}
\begin{tabular}{lccc}
\toprule
Model & ESOL & FreeSolv & Lipophilicity \\
\midrule
RF (Morgan, $r=2$) & \sd{1.340}{0.352} & \sd{0.996}{0.294} & \sd{0.832}{0.032} \\
D-MPNN (Chemprop) & \sd{0.914}{0.051} & \sd{0.480}{0.133} & \sd{0.622}{0.102} \\
GCN & \sd{0.789}{0.158} & \sd{0.521}{0.191} & \sd{0.650}{0.053} \\
GAT & \sd{0.794}{0.111} & \sd{0.477}{0.081} & \sd{0.591}{0.032} \\
GIN & \sdsecond{0.746}{0.095} & \sd{0.468}{0.131} & \sd{0.593}{0.062} \\
EquiformerV2 & \sd{0.759}{0.063} & \sdbest{0.459}{0.103} & \sd{0.654}{0.038} \\
MACE & \sd{0.821}{0.063} & \sdsecond{0.462}{0.077} & \sd{0.725}{0.100} \\
KA-GCN & \sd{0.760}{0.098} & \sd{0.510}{0.086} & \sd{0.614}{0.054} \\
PRISM & \sdbest{0.741}{0.057} & \sdbest{0.459}{0.086} & \sd{0.628}{0.031} \\
\midrule
\multicolumn{4}{l}{\emph{Pre-trained bank (Hu et al.\cite{hu2020strategies})}} \\
GIN, pre-trained (last layer) & \sd{0.932}{0.141} & \sd{0.554}{0.172} & \sd{0.598}{0.042} \\
GIN, pre-trained (JK attention) & \sd{0.793}{0.139} & \sd{0.629}{0.250} & \sdsecond{0.586}{0.032} \\
PRISM, pre-trained bank & \sd{0.821}{0.137} & \sd{0.537}{0.037} & \sdbest{0.581}{0.030} \\
PRISM, pre-trained bank + fp bank$^{a}$ & \sd{0.726}{0.084} & \sd{0.478}{0.044} & \sd{0.636}{0.025} \\
\bottomrule
\end{tabular}}
\par\smallskip{\scriptsize\linespread{1.0}\selectfont\raggedright Test RMSE (mean $\pm$ SD over five shared 80/10/10 Bemis-Murcko scaffold partitions, lower is better) in $\log(\mathrm{mol/L})$ for ESOL, dataset SD units ($1\text{ unit}=3.84\text{\,kcal/mol}$) for FreeSolv, and $\log D$ for Lipophilicity. Bold and underline mark the best and second-best ranked values per column. $^{a}$Unselected variant with the fingerprint-resolved bank (Table~\ref{tab:pretrained}), not ranked.\par}
\end{table}

The spread across partitions is as large as the differences among the leading models. The SD over the five scaffold partitions is 0.06 to 0.16 on ESOL and 0.08 to 0.13 on FreeSolv for the five best models, compared with a margin of 0.005 between PRISM and GIN on ESOL and a difference of $2\times10^{-4}$ between PRISM and EquiformerV2 on FreeSolv. On the individual partitions (Supporting Information Table~S9), PRISM is better than the baseline with the best mean on three of five ESOL and three of five FreeSolv partitions, two of five on BACE, and one of five on BBBP, and it is the best of all models on two BACE, one ESOL, and one BBBP partition. It does not beat the leading model on any Lipophilicity or SIDER partition. PRISM has the lowest mean error on ESOL, ties for lowest on FreeSolv, and trails GAT on Lipophilicity.

Table~\ref{tab:classification} reports the classification endpoints BACE, BBBP, and SIDER under scaffold partitions for the same models. PRISM is second on BACE ($0.898\pm0.026$ vs $0.904\pm0.038$ for the random forest) and on BBBP ($0.885\pm0.037$ vs $0.889\pm0.040$ for D-MPNN), within 0.006 and 0.004 of the best, and no baseline is significantly better than PRISM on either. On SIDER, the random forest ($0.639\pm0.028$) is significantly better than PRISM ($0.609\pm0.033$), and the pre-trained GIN readouts are also higher, consistent with an endpoint of 27 sparse side-effect labels on 1427 molecules, where exact substructure counts are a strong prior.

\begin{table}[tb]
\centering
\caption{Classification under scaffold partitions.}
\label{tab:classification}
{\small\setlength{\tabcolsep}{4pt}
\begin{tabular}{lccc}
\toprule
Model & BACE & BBBP & SIDER \\
\midrule
RF (Morgan, $r=2$) & \sdbest{0.904}{0.038} & \sd{0.878}{0.032} & \sdsecond{0.639}{0.028} \\
D-MPNN (Chemprop) & \sd{0.891}{0.039} & \sdbest{0.889}{0.040} & \sd{0.594}{0.025} \\
GCN & \sd{0.854}{0.061} & \sd{0.855}{0.041} & \sd{0.610}{0.017} \\
GAT & \sd{0.848}{0.042} & \sd{0.880}{0.041} & \sd{0.620}{0.020} \\
GIN & \sd{0.841}{0.025} & \sd{0.864}{0.041} & \sd{0.621}{0.027} \\
EquiformerV2 & \sd{0.850}{0.071} & \sd{0.840}{0.040} & \sd{0.577}{0.034} \\
MACE & \sd{0.853}{0.067} & \sd{0.872}{0.029} & \sd{0.610}{0.025} \\
KA-GCN & \sd{0.891}{0.037} & \sd{0.881}{0.034} & \sd{0.589}{0.036} \\
PRISM & \sdsecond{0.898}{0.026} & \sdsecond{0.885}{0.037} & \sd{0.609}{0.033} \\
\midrule
\multicolumn{4}{l}{\emph{Pre-trained bank (Hu et al.\cite{hu2020strategies})}} \\
GIN, pre-trained (last layer) & \sd{0.884}{0.066} & \sd{0.869}{0.030} & \sd{0.636}{0.029} \\
GIN, pre-trained (JK attention) & \sd{0.897}{0.041} & \sd{0.868}{0.059} & \sdbest{0.641}{0.028} \\
PRISM, pre-trained bank & \sd{0.888}{0.034} & \sd{0.865}{0.057} & \sd{0.625}{0.031} \\
PRISM, pre-trained bank + fp bank$^{a}$ & \sd{0.905}{0.035} & \sd{0.882}{0.018} & \sd{0.610}{0.023} \\
\bottomrule
\end{tabular}}
\par\smallskip{\scriptsize\linespread{1.0}\selectfont\raggedright Test ROC-AUC of the predicted positive-class probability (mean $\pm$ SD over five shared scaffold partitions, higher is better), single-task for BACE and BBBP and averaged over 27 tasks for SIDER. Bold and underline mark the best and second-best ranked values per column. $^{a}$Unselected variant with the fingerprint-resolved bank (Table~\ref{tab:pretrained}), not ranked.\par}
\end{table}

No model is best everywhere. Across the six endpoints the best mean score belongs to PRISM (ESOL), PRISM and EquiformerV2 (FreeSolv), GAT (Lipophilicity), the random forest (BACE and SIDER), and D-MPNN (BBBP), and PRISM is first or second on four. Summarized over endpoints, no baseline is significantly better than PRISM on four of six (ESOL, FreeSolv, BACE, BBBP), while only GAT on Lipophilicity and the random forest on SIDER are significantly better than PRISM (Supporting Information Table~S8). Two considerations qualify this comparison. First, it compares complete architectures, since PRISM differs from the conventional backbones in regularization, readout, and capacity as well as in allocation. The next two sections address capacity directly and isolate the allocation module through ablations. Second, every model, PRISM included, was selected on a single validation split of 64 to 420 molecules, and differences smaller than one partition SD are not resolved by five partitions.

\subsection{Parameter Count, Computational Cost, and Tuning Budget}
\label{sec:capacity}
Table~\ref{tab:cost} reports, for every model, the trainable parameter count, FLOPs per molecule, training time per epoch, and the number of tuning trials, which is the same for all models. PRISM has 11.4~M parameters (5.6 times the 2.0~M of GIN), of which 7.3~M form the embedding table of the fingerprint-resolved bank, a lookup that adds memory but almost no computation. It needs 1.7 times the FLOPs per molecule (118.5 vs 70.3~MFLOPs) and 2.5 times the time per epoch of GIN (0.72 vs 0.29~s). Additional protein-ligand affinity results and measured parameter, time, and memory costs on PDBbind against four geometric baselines are reported in Supporting Information Section~S6 (Table~S7). To separate capacity from design, the last two rows of Table~\ref{tab:cost} widen GIN to the parameter count of PRISM and narrow PRISM to that of GIN. Widening GIN to 11.5~M parameters degrades Lipophilicity ($0.661\pm0.046$ vs $0.593\pm0.062$) and is unstable across runs on ESOL ($0.862\pm0.175$ vs $0.746\pm0.095$, with a second independent five-seed run at identical settings giving $0.740\pm0.128$, matching PRISM's $0.741\pm0.057$), whereas narrowing PRISM to 2.0~M parameters raises its error on both endpoints (ESOL $0.782\pm0.134$ vs $0.741\pm0.057$, and Lipophilicity, at the smallest width the fingerprint table allows, $0.759\pm0.038$ vs $0.628\pm0.031$). Capacity therefore contributes to PRISM's results, while on Lipophilicity added capacity alone in GIN does not reproduce them.

\begin{table}[tb]
\centering
\caption{Model size, computational cost, and tuning budget, with a parameter-matched control.}
\label{tab:cost}
{\footnotesize\setlength{\tabcolsep}{1.3pt}
\begin{tabular}{lcccccc}
\toprule
Model & \shortstack{Params\\(M)} & \shortstack{MFLOPs\\per molecule} & \shortstack{Time per\\epoch (s)} & Trials & ESOL & Lipophilicity \\
\midrule
RF (Morgan, $r=2$) & N/A & N/A & 0.72$^{a}$ & 10 & \sd{1.340}{0.352} & \sd{0.832}{0.032} \\
D-MPNN (Chemprop) & 1.51 & 129.8 & 0.61 & 10 & \sd{0.914}{0.051} & \sd{0.622}{0.102} \\
GCN & 1.46 & 16.5 & 0.31 & 10 & \sd{0.789}{0.158} & \sd{0.650}{0.053} \\
GAT & 1.06 & 47.8 & 0.27 & 10 & \sd{0.794}{0.111} & \sd{0.591}{0.032} \\
GIN & 2.05 & 70.3 & 0.29 & 10 & \sdsecond{0.746}{0.095} & \sd{0.593}{0.062} \\
EquiformerV2 & 26.66 & 16993.8 & 3.83 & 10 & \sd{0.759}{0.063} & \sd{0.654}{0.038} \\
MACE & 38.78 & 35081.8 & 5.41 & 10 & \sd{0.821}{0.063} & \sd{0.725}{0.100} \\
KA-GCN & 3.89 & 99.9 & 0.35 & 10 & \sd{0.760}{0.098} & \sd{0.614}{0.054} \\
PRISM & 11.43 & 118.5 & 0.72 & 10 & \sdbest{0.741}{0.057} & \sd{0.628}{0.031} \\
\midrule
\multicolumn{7}{l}{\emph{Pre-trained bank (Hu et al.\cite{hu2020strategies})}} \\
GIN, pre-trained (last layer) & 2.40 & 62.8 & 0.17 & 10 & \sd{0.932}{0.141} & \sd{0.598}{0.042} \\
GIN, pre-trained (JK attention) & 2.40 & 62.9 & 0.32 & 10 & \sd{0.793}{0.139} & \sdsecond{0.586}{0.032} \\
PRISM, pre-trained bank & 2.45 & 64.4 & 0.18 & 10 & \sd{0.821}{0.137} & \sdbest{0.581}{0.030} \\
PRISM, pre-trained bank + fp bank$^{b}$ & 11.05 & 64.4 & 0.19 & 10 & \sd{0.726}{0.084} & \sd{0.636}{0.025} \\
\midrule
\multicolumn{7}{l}{\emph{Parameter-matched control}} \\
GIN, widened to PRISM size & 11.49 & 396.2 & 0.31 & N/A & \sd{0.862}{0.175}$^{c}$ & \sd{0.661}{0.046} \\
PRISM, narrowed to GIN size & 2.01 & 6.5 & 0.68 & N/A & \sd{0.782}{0.134} & \sd{0.759}{0.038} \\
\bottomrule
\end{tabular}}
\par\smallskip{\scriptsize\linespread{1.0}\selectfont\raggedright Trainable parameters, forward-pass MFLOPs per molecule (ESOL test-set mean), and training time per epoch on ESOL (one NVIDIA A100-SXM4-80GB GPU). ESOL and Lipophilicity report test RMSE (mean $\pm$ SD, five scaffold partitions). Parameter-matched controls adjust hidden width only. $^{a}$CPU fit time on 16 threads. $^{b}$Unselected variant with the fingerprint-resolved bank (Table~\ref{tab:pretrained}), not ranked. $^{c}$A second independent five-seed run with identical settings on ESOL gave $0.740\pm0.128$.\par}
\end{table}

\subsection{Component Ablations and Node-Dependent Layer Weighting}
Each ablation arm changes one component of the final model, retrains it with the setting selected for the dataset, and is compared with the full model on the same five partitions (paired $t$-tests over partitions). Table~\ref{tab:ablation} tabulates the component removals on SIDER, and Supporting Information Table~S10 (Section~S8) reports every arm on all six endpoints.

\begin{table}[tb]
\centering
\caption{Absolute ablation results on SIDER: component removals.}
\label{tab:ablation}
{\scriptsize\setlength{\tabcolsep}{3.5pt}
\begin{tabular}{lcccccc}
\toprule
 & \shortstack{PRISM\\(full)} & \shortstack{Without\\$\mathcal L_{\mathrm{feat}}$} & \shortstack{With\\$\mathcal L_{\kappa}$} & \shortstack{Mean pooling\\only} & \shortstack{Sum pooling\\only} & \shortstack{No adaptive\\allocation} \\
\midrule
SIDER ROC-AUC $\uparrow$ & \sdbest{0.615}{0.032} & \sd{0.602}{0.035} & \sd{0.597}{0.014} & \sd{0.589}{0.027} & \sd{0.602}{0.024} & \sd{0.601}{0.026} \\
\bottomrule
\end{tabular}}
\par\smallskip{\scriptsize\linespread{1.0}\selectfont\raggedright Mean $\pm$ SD over five runs: test ROC-AUC with a fixed output-to-label mapping under scaffold partitions (higher is better). All arms share the partition, features, optimizer, schedule, readout, and evaluation code with the full model and differ only in the stated component. All arms, including the full-model reference, are independent five-seed runs, so the full-model value differs slightly from the PRISM row of Table~\ref{tab:classification}. The same arms and the node-dependent layer-weighting alternatives on all six endpoints are given in Supporting Information Table~S10.\par}
\end{table}

Removing adaptive allocation significantly lowers ROC-AUC on SIDER ($0.601$ vs $0.615$) and leaves the other five endpoints within the partition spread (Supporting Information Table~S10). Including the scalar-field penalty $\mathcal L_\kappa$, or removing the feature-energy hinge, changes no endpoint significantly except for a lower SIDER ROC-AUC when the hinge is removed ($0.602$).

Sum pooling is the only component whose removal is costly on most endpoints. With mean pooling alone, RMSE rises from $0.611$ to $0.690$ on ESOL ($+13\%$) and from $0.254$ to $0.316$ on FreeSolv ($+24\%$), and ROC-AUC falls from $0.615$ to $0.589$ on SIDER, all significantly, while Lipophilicity is unchanged ($0.611$ vs $0.607$) and sum pooling alone matches the full model. The ablation shows that the accumulated magnitude retained by sum pooling matters for these targets. One possible explanation is molecular size: mean pooling discards the atom count, and solubility and hydration free energy depend more directly on size and solvent-exposed surface than $\log D$ does, which is consistent with the absence of an effect on Lipophilicity.

The node-dependent arms answer whether a less constrained node-wise scheme does better under the same encoder, bank, local branch, readout, training, and parameter count. None is significantly better than PRISM on any endpoint (Supporting Information Table~S10). The free node-wise layer gate, which uses the same input as PRISM but outputs $L$ unconstrained logits, is significantly worse on FreeSolv ($0.273$ vs $0.254$) and within the partition spread elsewhere. Node-wise Jumping Knowledge attention is significantly worse on BACE ($0.890$ vs $0.899$), SIDER ($0.594$), and Lipophilicity ($0.631$ vs $0.607$). Jumping Knowledge concatenation and max pooling, as well as PRISM-env (which routes from $h^{(1)}$ on every endpoint, so on BBBP and SIDER, where validation already selected $h^{(1)}$ routing, the full model and PRISM-env are two independent five-seed runs of that configuration), are within the partition spread of the full model on every endpoint. The one-parameter profile is at least as accurate as unconstrained node-wise schemes while remaining analyzable in closed form, and removing the allocation lowers accuracy significantly on SIDER while leaving the other five endpoints within the partition spread.

Robustness to input perturbation was measured for the final model. With Gaussian noise of SD up to 0.20 added to the encoded atom features at training and evaluation time, no endpoint changes beyond the spread across partitions. BACE ROC-AUC is 0.899 without noise and 0.909 at 0.20, BBBP 0.870 and 0.857, ESOL RMSE 0.611 and 0.626, and FreeSolv RMSE 0.254 and 0.247 (Supporting Information Table~S4), with no monotone trend on any endpoint. The ablation arms were not run under noise, so this measures the sensitivity of the full model only.

\subsection{Allocation on a Pre-Trained Message-Passing Stack}
\label{sec:pretrained}
A pre-trained network is a bank whose layers were shaped by millions of molecules rather than by the few hundred to few thousand of each benchmark, and it is the setting in which reading only the last layer is the default practice. Table~\ref{tab:pretrained} compares four readouts of the pre-trained GIN of Hu et al.\cite{hu2020strategies} under identical fine-tuning. PRISM allocation improves on the standard last-layer readout on four of six endpoints, most on ESOL ($0.821\pm0.137$ vs $0.932\pm0.141$), with smaller gains on FreeSolv ($0.537$ vs $0.554$), Lipophilicity ($0.581$ vs $0.598$), and BACE ($0.888$ vs $0.884$), and is slightly worse on BBBP and SIDER (by 0.004 and 0.011). Among the three readouts of the unaugmented pre-trained bank, PRISM allocation is the best readout on FreeSolv and Lipophilicity and ranks second on ESOL and BACE, where node-wise Jumping Knowledge attention is best. On Lipophilicity its RMSE of $0.581\pm0.030$ is the lowest of every model in Table~\ref{tab:scaffold}. The differences from the last layer are not significant over five partitions except for the fingerprint-resolved variant on ESOL ($p=0.030$), which was not selected on validation. That variant, which adds the fingerprint-resolved bank to the pre-trained stack, has lower error than the selected one on ESOL ($0.726\pm0.084$) and BACE ($0.905\pm0.035$) but higher error on Lipophilicity ($0.636\pm0.025$) and SIDER ($0.610\pm0.023$, significantly below the last layer). Validation preferred the variant without the fingerprint-resolved bank on five of six endpoints, so that variant is used as the primary pre-trained PRISM model in all rankings. On a stack whose intermediate layers already encode chemistry at several radii, letting each atom choose which layer to read is more useful than reading the last, although it does not dominate the less constrained node-wise attention.

\begin{table}[tb]
\centering
\caption{PRISM allocation on a pre-trained message-passing stack.}
\label{tab:pretrained}
{\small\setlength{\tabcolsep}{4pt}
\begin{tabular}{lcccc}
\toprule
 & \multicolumn{4}{c}{Readout of the pre-trained GIN} \\
\cmidrule(lr){2-5}
Endpoint & Last layer & JK attention & PRISM$^{a}$ & PRISM + fp bank \\
\midrule
ESOL (RMSE $\downarrow$) & \sd{0.932}{0.141} & \sd{0.793}{0.139} & \sd{0.821}{0.137} & \sdbest{0.726}{0.084} \\
FreeSolv (RMSE $\downarrow$) & \sd{0.554}{0.172} & \sd{0.629}{0.250} & \sd{0.537}{0.037} & \sdbest{0.478}{0.044} \\
Lipophilicity (RMSE $\downarrow$) & \sd{0.598}{0.042} & \sd{0.586}{0.032} & \sdbest{0.581}{0.030} & \sd{0.636}{0.025} \\
BACE (ROC-AUC $\uparrow$) & \sd{0.884}{0.066} & \sd{0.897}{0.041} & \sd{0.888}{0.034} & \sdbest{0.905}{0.035} \\
BBBP (ROC-AUC $\uparrow$) & \sd{0.869}{0.030} & \sd{0.868}{0.059} & \sd{0.865}{0.057} & \sdbest{0.882}{0.018} \\
SIDER (ROC-AUC $\uparrow$) & \sd{0.636}{0.029} & \sdbest{0.641}{0.028} & \sd{0.625}{0.031} & \sd{0.610}{0.023} \\
\bottomrule
\end{tabular}}
\par\smallskip{\scriptsize\linespread{1.0}\selectfont\raggedright Mean $\pm$ SD over the five scaffold partitions of Tables~\ref{tab:scaffold} and \ref{tab:classification} (best per row in bold). All columns fine-tune the five-layer pre-trained GIN of Hu et al.\cite{hu2020strategies} under identical 10-trial tuning and readout, differing only in layer combination (JK: node-wise Jumping Knowledge, fp bank: eq~\ref{eq:fpbank}). $^{a}$Selected by mean validation rank across endpoints (1.17 vs 1.83) for Tables~\ref{tab:scaffold} and \ref{tab:classification}.\par}
\end{table}

\subsection{Transfer Across Backbones, Profile Permutation, and the ADR Benchmark}
\label{sec:assignment}
The ablations above change the architecture and retrain it. Three further experiments test more directly whether the per-atom allocation itself drives performance.

First, the layer-combination rule alone was attached to three standard backbones, GCN, GIN, and GAT, without the local branch or the regularizers, and compared with the same backbone reading its last layer and with node-wise Jumping Knowledge (Table~\ref{tab:backbones}, where all three rules use the linear atom encoder with the validation-selected depth and width for each backbone and endpoint). Within each backbone the PRISM allocation gives the best value on 7 of the 12 backbone-endpoint pairs, the last layer on 3, and node-wise Jumping Knowledge on 3 (with one tie counted for both). The clearest gain is on FreeSolv, where the allocation lowers RMSE relative to the last layer by 29\% for GCN ($0.673$ vs $0.951$) and 12\% for GIN ($0.596$ vs $0.676$) and matches it for GAT. On Lipophilicity and BACE the three rules differ by at most 0.022, and on ESOL the allocation is worse than the last layer for GIN ($0.819$ vs $0.746$). The allocation transfers as a layer-combination rule and helps most on FreeSolv, although the effect is modest and nonuniform, with most differences lying within one partition SD.

\begin{table}[tb]
\centering
\caption{Transfer of the allocation module to standard message-passing backbones.}
\label{tab:backbones}
{\footnotesize\setlength{\tabcolsep}{3pt}
\begin{tabular}{llcccc}
\toprule
Backbone & Layer combination & ESOL $\downarrow$ & FreeSolv $\downarrow$ & Lipo $\downarrow$ & BACE $\uparrow$ \\
\midrule
GCN & last layer & \sdbest{0.842}{0.150} & \sd{0.951}{0.404} & \sd{0.673}{0.070} & \sd{0.854}{0.061} \\
GCN & JK attention & \sd{0.895}{0.172} & \sd{0.931}{0.354} & \sd{0.665}{0.065} & \sdbest{0.859}{0.046} \\
GCN & PRISM & \sdbest{0.842}{0.165} & \sdbest{0.673}{0.344} & \sdbest{0.664}{0.056} & \sd{0.837}{0.034} \\
\midrule
GIN & last layer & \sdbest{0.746}{0.095} & \sd{0.676}{0.108} & \sd{0.650}{0.040} & \sd{0.841}{0.025} \\
GIN & JK attention & \sd{0.873}{0.241} & \sd{0.648}{0.120} & \sd{0.658}{0.028} & \sd{0.838}{0.075} \\
GIN & PRISM & \sd{0.819}{0.185} & \sdbest{0.596}{0.079} & \sdbest{0.647}{0.045} & \sdbest{0.844}{0.041} \\
\midrule
GAT & last layer & \sd{0.794}{0.111} & \sd{0.595}{0.099} & \sdbest{0.619}{0.031} & \sd{0.863}{0.052} \\
GAT & JK attention & \sdbest{0.771}{0.040} & \sdbest{0.560}{0.133} & \sd{0.638}{0.044} & \sd{0.854}{0.049} \\
GAT & PRISM & \sd{0.779}{0.117} & \sd{0.594}{0.193} & \sd{0.634}{0.059} & \sdbest{0.864}{0.045} \\
\bottomrule
\end{tabular}}
\par\smallskip{\scriptsize\linespread{1.0}\selectfont\raggedright Test RMSE (ESOL, FreeSolv, Lipophilicity [Lipo]) and ROC-AUC (BACE), mean $\pm$ SD over five scaffold partitions (best per backbone block in bold). Within each block, models share architecture and training and differ only in layer combination: last layer, node-wise Jumping Knowledge (JK) attention, or PRISM allocation (eqs~\ref{eq:center} to \ref{eq:aggregate}, without the local branch or regularizers). All models in this table use the linear atom encoder with the validation-selected depth and width for each backbone and endpoint, whereas Tables~\ref{tab:scaffold} and \ref{tab:classification} report whichever encoder (linear or categorical embedding) validation preferred.\par}
\end{table}

Second, in trained PRISM models the layer-weight profiles were permuted among the atoms of each molecule, which keeps the molecule's multiset of profiles and its mean allocated depth but breaks the correspondence between atoms and profiles, and were alternatively replaced by the molecule-average profile (Table~\ref{tab:shuffle}). In the final model, permutation raises Lipophilicity RMSE by $0.107\pm0.035$ and lowers SIDER ROC-AUC by $0.024\pm0.017$, both significant over the five models ($p=0.002$ and $0.030$), raises ESOL and FreeSolv RMSE by $0.020\pm0.023$ and $0.014\pm0.015$ (not significant), and changes BBBP and BACE ROC-AUC by at most $0.003$ ($+0.0002\pm0.0002$ on BACE before rounding). On BBBP 76.2\% of atoms lie on a clipping plateau, which removes center variation among atoms clipped at the same boundary, while the BACE configuration fuses a Morgan fingerprint into the readout (Supporting Information Table~S1) and has 46\% to 61\% of atoms on the plateau in three of the five models. Replacing every profile by the molecule mean has smaller effects of the same sign. On the pre-trained bank, permutation degrades performance on every endpoint, raising RMSE by $0.140\pm0.085$, $0.109\pm0.049$, and $0.219\pm0.091$ on ESOL, FreeSolv, and Lipophilicity and lowering ROC-AUC by 0.016, 0.026, and 0.014 on BACE, BBBP, and SIDER. Because the network, bank, and routing outputs are unchanged, a deterioration under permutation can only come from which atom receives which profile. Together with the fraction of atoms on the clipping plateau reported in the same table, this experiment tests atom-level heterogeneity directly.

\begin{table}[tb]
\centering
\caption{Effect of disrupting the atom-profile correspondence in trained models.}
\label{tab:shuffle}
{\scriptsize\setlength{\tabcolsep}{1.2pt}
\begin{tabular}{lcccccc}
\toprule
Perturbation & ESOL & FreeSolv & Lipo & BACE & BBBP & SIDER \\
\midrule
Permuted $p_i$ & \sd{+0.020}{0.023} & \sd{+0.014}{0.015} & \sd{+0.107}{0.035} & \sd{+0.000}{0.000} & \sd{-0.003}{0.007} & \sd{-0.024}{0.017} \\
Molecule mean $\bar p$ & \sd{+0.018}{0.021} & \sd{+0.017}{0.020} & \sd{+0.063}{0.033} & \sd{+0.000}{0.000} & \sd{-0.002}{0.006} & \sd{-0.017}{0.016} \\
\midrule
\multicolumn{7}{l}{\emph{PRISM allocation on the pre-trained bank (Table~\ref{tab:pretrained})}} \\
Pre-trained, permuted $p_i$ & \sd{+0.140}{0.085} & \sd{+0.109}{0.049} & \sd{+0.219}{0.091} & \sd{-0.016}{0.027} & \sd{-0.026}{0.016} & \sd{-0.014}{0.013} \\
Pre-trained, mean $\bar p$ & \sd{+0.099}{0.081} & \sd{+0.069}{0.046} & \sd{+0.131}{0.068} & \sd{-0.006}{0.016} & \sd{-0.011}{0.007} & \sd{-0.009}{0.013} \\
\midrule
On plateau (\%, full model) & 13.2 & 39.6 & 2.5 & 31.5 & 76.2 & 3.5 \\
\bottomrule
\end{tabular}}
\par\smallskip{\scriptsize\linespread{1.0}\selectfont\raggedright $\Delta$RMSE (ESOL, FreeSolv, Lipophilicity [Lipo], random partitions) and $\Delta$ROC-AUC (BACE, BBBP, SIDER, scaffold partitions) relative to the unperturbed model (mean $\pm$ SD over the five full-model runs of Supporting Information Table~S10, 20 permutations each, where $+\Delta$RMSE and $-\Delta$ROC-AUC indicate degradation). With trained weights fixed, ``Permuted $p_i$'' shuffles profiles within each molecule, ``Molecule mean $\bar p$'' assigns the molecular average, and ``On plateau'' is the percentage of test atoms with clipped centers (on BACE, both full-model entries are $+0.0002$ before rounding).\par}
\end{table}

Third, on the synthetic ADR benchmark of eq~\ref{eq:adr}, where heteroatoms must be read at one hop and carbons at $r_C$ hops, PRISM reaches test MAE $0.757\pm0.240$ at $r_C=2$ and $1.052\pm0.133$ at $r_C=3$, within 0.022 of the best scheme (the node-wise layer gate at $r_C=2$, node-wise Jumping Knowledge at $r_C=3$), and it has the highest MAE at $r_C=4$ ($1.400\pm0.420$ vs $1.169\pm0.138$ for node-wise Jumping Knowledge). Every scheme reaches an MAE of about 1 on targets of SD near 30, because a residual GIN that reads its last layer already represents both radii. Permuting PRISM's profiles among the atoms of each molecule raises its MAE to 2.37, 2.58, and 2.53 at $r_C=2,3,4$, up to three times the unperturbed error, so the trained network relies on which atom receives which profile even where that assignment does not translate into a lower error than the alternatives (Table~\ref{tab:adr}). The same holds for PRISM on the pre-trained bank, whose MAE rises from the range 0.95 to 1.53 up to the range 2.14 to 3.37 under permutation. The final model of Tables~\ref{tab:scaffold} and \ref{tab:classification}, whose fingerprint-resolved bank supplies radius-$r$ environment identifiers closely related to the counts the target is built from, reaches MAE $0.735\pm0.050$, $1.238\pm0.089$, and $1.521\pm0.086$, the lowest of all rows at $r_C=2$, and permutation raises its MAE to $0.965$, $1.366$, and $1.592$. The effect is smaller than on the plain GIN bank, probably because most atoms lie on a clipping plateau in several of its runs. This benchmark shows that the mechanism can exploit atom-type-conditioned structure when it exists.

\begin{table}[tb]
\centering
\caption{Synthetic atom-dependent-radius (ADR) benchmark.}
\label{tab:adr}
{\footnotesize\setlength{\tabcolsep}{4pt}
\begin{tabular}{lccc}
\toprule
Model & $r_C=2$ & $r_C=3$ & $r_C=4$ \\
\midrule
\multicolumn{4}{l}{\emph{Same six-layer GIN bank, readout differs}} \\
GIN, last layer & \sd{0.785}{0.133} & \sd{1.087}{0.099} & \sd{1.187}{0.154} \\
GIN, uniform average & \sd{0.742}{0.099} & \sd{1.066}{0.196} & \sd{1.263}{0.183} \\
GIN, shared layer attention & \sd{1.007}{0.263} & \sd{1.075}{0.097} & \sdsecond{1.184}{0.161} \\
Node-wise layer gate & \sdsecond{0.738}{0.182} & \sd{1.163}{0.170} & \sd{1.265}{0.124} \\
JK, node-wise attention & \sd{0.830}{0.188} & \sdbest{1.030}{0.114} & \sdbest{1.169}{0.138} \\
PRISM allocation & \sd{0.757}{0.240} & \sdsecond{1.052}{0.133} & \sd{1.400}{0.420} \\
\quad permuted within molecule (GIN bank) & \sd{2.370}{1.091} & \sd{2.576}{1.074} & \sd{2.533}{1.433} \\
\midrule
\multicolumn{4}{l}{\emph{PRISM as in Tables~\ref{tab:scaffold} and \ref{tab:classification}}} \\
PRISM & \sdbest{0.735}{0.050} & \sd{1.238}{0.089} & \sd{1.521}{0.086} \\
\quad permuted within molecule (PRISM) & \sd{0.965}{0.454} & \sd{1.366}{0.247} & \sd{1.592}{0.088} \\
\midrule
\multicolumn{4}{l}{\emph{Pre-trained GIN of Hu et al.\cite{hu2020strategies}}} \\
Pre-trained, last layer & \sd{1.249}{0.084} & \sd{1.665}{0.059} & \sd{1.769}{0.079} \\
Pre-trained, JK attention & \sd{1.027}{0.031} & \sd{1.430}{0.069} & \sd{1.585}{0.129} \\
Pre-trained, PRISM & \sd{0.953}{0.078} & \sd{1.455}{0.138} & \sd{1.527}{0.074} \\
\quad permuted within molecule (pre-trained) & \sd{2.135}{0.716} & \sd{3.370}{0.419} & \sd{2.765}{0.753} \\
Pre-trained, PRISM + fp bank$^{a}$ & \sd{1.018}{0.210} & \sd{1.403}{0.167} & \sd{1.680}{0.184} \\
\quad permuted within molecule (+ fp bank)$^{a}$ & \sd{2.224}{0.540} & \sd{2.668}{0.177} & \sd{3.673}{0.665} \\
\bottomrule
\end{tabular}}
\par\smallskip{\scriptsize\linespread{1.0}\selectfont\raggedright Test MAE on eq~\ref{eq:adr} (mean $\pm$ SD, five random partitions, lower is better). Bold and underline mark the best and second-best ranked rows. The first block shares a six-layer GIN bank and differs only in layer combination. ``Permuted within molecule'' shuffles $p_i$ within each molecule (20 permutations, not ranked). $^{a}$Unselected variant (Table~\ref{tab:pretrained}, not ranked). Both PRISM and this variant include radius-$r$ fingerprint embeddings (eq~\ref{eq:fpbank}).\par}
\end{table}

\subsection{Depth Sweep on Peptides-struct: Accuracy, Time, and Memory}
\label{sec:cost}
\begin{figure}[!htbp]
\centering
\includegraphics[width=\textwidth]{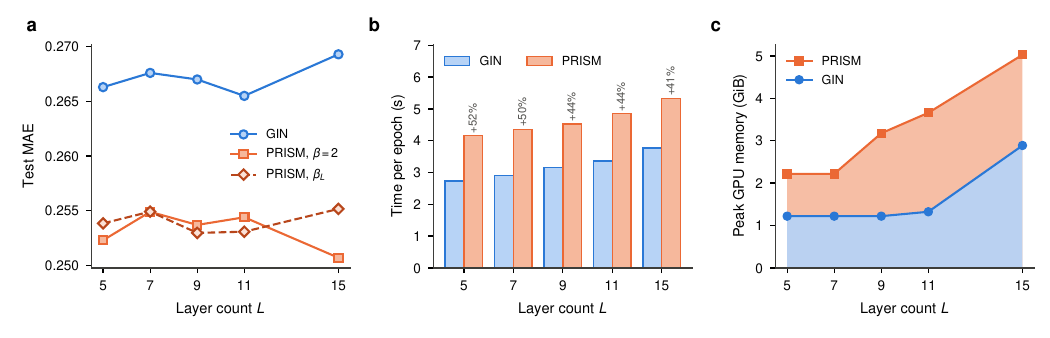}
\caption{Peptides-struct depth sweep. (a) Test MAE against layer count for GIN and PRISM with $\beta=2$ (Supporting Information Table~S2) and PRISM with the depth-scaled $\beta_L=(L-1)/2$ (dashed, Supporting Information Table~S3), (b) time per epoch, with the relative overhead of PRISM, and (c) peak GPU memory in GiB (1\,GiB = 1024\,MiB, with values in MiB in Supporting Information Table~S2). Each point is a single run per configuration, and lines connect tested configurations.}
\label{fig:depth}
\end{figure}

Peptides-struct tests the regime of large graphs in which information must travel far. Here the accuracy benefit and its computational cost are measured together (Figure~\ref{fig:depth}, with full values in Supporting Information Section S4).

PRISM has lower MAE than GIN\cite{xu2018powerful} at all five layer counts, from 0.2523 vs 0.2663 at five layers to 0.2507 vs 0.2693 at 15 layers, the latter a 6.9\% reduction. Within PRISM, error does not fall monotonically with available depth and moves between 0.2507 and 0.2549 across the sweep, and the 15-layer configuration is better than the 11-layer one by 0.0037, a difference single runs cannot resolve. PRISM is consistently ahead of GIN at matched depth, and adding layers within PRISM does not reliably help.

The computational overhead is substantial. Relative to GIN, PRISM needs 41\% to 52\% more time per epoch and 74\% to 176\% more peak memory across the sweep, with the smallest memory overhead at 15 layers (5146 vs 2956~MiB, or 5.03 vs 2.89~GiB) and the largest at 11 layers (3746 vs 1356~MiB, or 3.66 vs 1.32~GiB). This overhead follows from computing every bank state. Because every bank state is evaluated before any is weighted, shifting an atom's allocation toward earlier states saves no computation: the mechanism organizes how intermediate states are combined at readout and is not an early-exit scheme. A deployment decision has to weigh a 4\% to 7\% error reduction against substantially higher memory. Two caveats apply to the accuracy comparison in this sweep. First, the allocation parameters were trained at a much lower learning rate and held fixed early in training (Methods), so this sweep is a weak test of learned allocation. Second, the GIN baseline here is weaker than tuned GINE results reported for this benchmark, so the comparison establishes the cost of the design more firmly than its accuracy advantage. With $\beta=2$ the unclipped centers cover at most four index units of the bank, which for nine or more layers is less than half of it (Methods). Running the sweep with the depth-scaled coefficient $\beta_L=(L-1)/2$ gives test MAE of 0.2538, 0.2549, 0.2530, 0.2531, and 0.2552 at 5, 7, 9, 11, and 15 layers, vs 0.2529, 0.2534, 0.2527, 0.2513, and 0.2529 when $\beta=2$ is rerun under the same script (Supporting Information Table~S3), and the learned allocations under both settings are summarized in Supporting Information Section~S5. Scaling $\beta$ with depth does not change the MAE trend and does not widen the occupied range of $d_i$. At $L=15$ the 5th to 95th percentiles of the allocated index are 8.89 to 8.91 with $\beta=2$ and 8.46 to 8.62 with $\beta_L$, so on Peptides-struct nearly every atom receives the same depth under both settings, and the reference center $b$, trained with a learning rate of $10^{-5}$, stays at its initialization (Supporting Information Table~S6).

\section{Discussion}
This work provides a constrained answer to a design question that molecular graph networks usually settle by default, namely how much of each propagation scale a given atom should contribute to a molecular prediction. PRISM answers it with one bounded scalar per atom that positions a unimodal distribution over a bank of retained states. That construction is not the introduction of multi-scale molecular representation, which circular and learned fingerprints, Jumping Knowledge aggregation, and adaptive propagation already established\cite{rogers2010ecfp,duvenaud2015fingerprints,xu2018jumping,spinelli2020adaptive}. The new element is the intermediate design point, where the profile is more flexible than a shared layer-weight vector and more restricted than free per-layer selection. Because a single coordinate translates the whole profile, its effect on allocated depth can be derived in closed form, including in the boundary regions where clipping saturates the response (Figure~\ref{fig:allocation}). At matched parameter count and under the same encoder, bank, readout, and training, neither the free node-wise gate nor node-wise Jumping Knowledge attention is significantly better than PRISM on any endpoint, the free node-wise gate is significantly worse on FreeSolv, and node-wise Jumping Knowledge attention is significantly worse on BACE, SIDER, and Lipophilicity (Supporting Information Table~S10). The restriction costs no measurable accuracy and provides a closed-form account of the allocation.

Under scaffold partitions with matched tuning and standard ROC-AUC, PRISM trained from scratch has the lowest or tied-lowest mean error on ESOL and FreeSolv, is second on BACE and BBBP within 0.006 of the best, and trails GAT on Lipophilicity and the fingerprint forest on SIDER. The spread across the five partitions is of the same size as the differences among the leading models, and no single baseline is best on more than two endpoints. PRISM is among the strongest models, though it does not dominate them. On the pre-trained bank (Table~\ref{tab:pretrained}), the operator, pre-training, fine-tuning, and trial settings are held fixed and only the rule that reads the layer states changes. There, PRISM allocation improves mean performance over the standard last-layer readout on four of six endpoints and reaches the lowest Lipophilicity error of all models. This is the setting the design targets directly: when intermediate layers already encode chemistry at several radii, letting each atom select which layers to read improves mean accuracy on most endpoints over reading only the last layer, although the per-endpoint differences lie within the five-partition spread. It is not uniformly better than node-wise Jumping Knowledge attention over the same stack, which is less constrained and improves over both the last layer and PRISM on three endpoints, and the trade-off between that flexibility and the analyzability of a one-parameter profile remains an open design question.

For molecular modeling, the ablations of the final model indicate where each component matters. Sum pooling is the only component whose removal is costly across multiple endpoints, with the largest losses on hydration free energy and solubility and none on lipophilicity, consistent with a size effect. Removing the learned allocation lowers accuracy on SIDER, and the unconstrained node-wise gate that replaces it is worse on FreeSolv. On the other endpoints removing the allocation leaves accuracy within the partition spread, so the allocation contributes to accuracy on some endpoints rather than all. The per-atom character of the allocation is supported separately. Within each trained model the allocated depth varies by about one bank index across atoms, mostly within molecules, and it differs systematically between atom types (Figure~\ref{fig:allocation}e, f, and g). Permuting profiles among the atoms of a molecule degrades performance significantly on Lipophilicity and SIDER, produces smaller shifts in the same direction on ESOL and FreeSolv, and raises the error on every regression endpoint when the bank is pre-trained (Table~\ref{tab:shuffle}). The rule also improves mean performance over the last-layer readout of a pre-trained GIN on four of six endpoints (Table~\ref{tab:pretrained}) and improves on the last-layer readout of GCN, GIN, and GAT backbones, most clearly on hydration free energy (Table~\ref{tab:backbones}).

Allocation does not skip computation. All $L$ bank states are evaluated before any are weighted, so an atom allocated toward early states contributes no saving, and the measured overhead on Peptides-struct is 41\% to 52\% in time per epoch and 74\% to 176\% in peak memory relative to GIN across the sweep, for error reductions of 4\% to 7\%. For a practitioner, the mechanism is worth its cost where accuracy on a long-range-dependent target dominates and memory is not the binding constraint, and hard to justify where a fingerprint model or a plain message-passing network is already adequate. Nothing in the design addresses over-squashing, and the heat-kernel ordering in Figure~\ref{fig:allocation}d concerns a linearized operator rather than the trained network with its residual updates, edge features, local branch, normalization, and readout\cite{alon2021bottleneck,rusch2023survey}.

The chemical interpretability of the routing scalar is limited. In the default configuration, the routing network sees an atom's encoded features before any neighbor aggregation, so the allocation it learns is a function of atom type and local descriptors, and atoms with identical feature vectors receive identical profiles in every molecular environment. In classical atomic-contribution and group-additivity models the atom's type likewise determines how much of its environment is relevant\cite{wildman1999crippen,benson1958additivity}, and PRISM can learn such a type-to-scale assignment from data. In the final models, allocated depth varies by atom type in a task-dependent way, with nitrogen allocated about two bank indices deeper than the model mean on FreeSolv and halogens about one index shallower on SIDER (Figure~\ref{fig:allocation}g). Default routing does not discover a dependence of useful propagation scale on the surrounding environment, and an association between $\kappa_i$ and aromaticity or charge is a re-expression of information already in the input. The environment-aware variant PRISM-env, which routes on $h_i^{(1)}$ (selected by validation on BBBP and SIDER), tests whether one-hop environment information matters on these endpoints, and PRISM-env is within the partition spread of the full model on every endpoint (Supporting Information Table~S10). Distinct scalar values inside a clipping plateau also produce identical layer weights, and the fraction of atoms on the plateau ranges from 0.9\% to 74.2\% across the scaffold partitions (Supporting Information Table~S5, and 2.5\% to 76.2\% in Table~\ref{tab:shuffle}). The permutation experiment shows that the assignment of profiles to atoms carries predictive information on several endpoints, significantly on Lipophilicity and SIDER. Whether the learned type-to-scale assignment corresponds to a physical length scale would require intervention on molecular structure and is left open.

\subsection{Limitations}
Neither scaffold nor random partitions reproduce prospective, time-ordered use, for which time splits are the more stringent test\cite{sheridan2013timesplit}. Parameter counts differ between tuned models, and only GIN has a parameter-matched control. The depth sweep is a single run per configuration, so its margin over GIN cannot be credited to the allocation alone. Removing the allocation lowers accuracy significantly only on SIDER among the six endpoints. Finally, the routing scalar of the default model depends only on the atom's own descriptors (except on BBBP and SIDER, where validation selected routing on $h^{(1)}$), so atoms with identical input descriptors receive identical profiles regardless of their wider molecular environment.

\section{Conclusions}
PRISM allocates propagation scale per atom by letting one bounded scalar position a unimodal distribution over a fingerprint-resolved bank of retained post-update graph states, paired with a regularized local branch and a mean-and-sum molecular readout. Because a single coordinate translates the profile at a given reference center and temperature, allocated depth is nondecreasing in the routing scalar, its sensitivity scales with the variance of the depth distribution, and it saturates once the center is clipped.

Under scaffold partitions with matched tuning, PRISM trained from scratch has the lowest or tied-lowest mean error among nine models on aqueous solubility and hydration free energy and is second on BACE and BBBP, with no baseline significantly better on these four endpoints. GAT is better on lipophilicity and the fingerprint random forest on SIDER. A parameter-matched control shows that capacity contributes to these results, while on lipophilicity added capacity alone does not reproduce them in GIN. Applied to a pre-trained GIN, PRISM allocation improves mean performance over last-layer readout on four of six endpoints and gives the lowest lipophilicity error overall. On Peptides-struct it reduces error relative to GIN at every tested depth at a cost of 41\% to 52\% more time and 74\% to 176\% more memory.

In the final model, sum pooling is the component whose removal is costly on most endpoints. Removing the learned allocation lowers accuracy on SIDER and leaves the other endpoints within the partition spread, while the node-dependent alternatives are not significantly better on any endpoint, and the parameter-matched node-wise gate and Jumping Knowledge attention are significantly worse on one and three endpoints, respectively. The allocation varies per atom with task-dependent differences between atom types, permuting learned profiles within molecules raises error significantly on lipophilicity and SIDER and on every regression endpoint of the pre-trained bank, and the allocation rule transfers to GCN, GIN, and GAT backbones, most clearly on hydration free energy. The rule acts as a learned, task-dependent counterpart of atom typing, whereas whether the learned type-to-scale assignment reflects a physical length scale of each property remains open.

\section*{Author Information}
\subsection*{Author Contributions}
R. Zhang designed the algorithm, implemented the code, conducted the experiments, and wrote the manuscript. H. Wen contributed to code implementation and experimental analysis. Z. Li assisted in figure preparation and manuscript polishing. J. Ouyang reviewed the manuscript. Z. Chu contributed to manuscript polishing and review. P. Li\`o contributed to idea discussion and algorithm design. X. Liu polished and refined the manuscript. S. Yiu contributed to idea discussion, manuscript polishing, and review. R. Zhang, H. Wen, and Z. Li contributed equally.
\subsection*{Notes}
The authors declare no competing financial interest.
\begin{acknowledgement}
This research is partly supported by the HKU-SCF FinTech Academy, the Shenzhen-Hong Kong-Macao Science and Technology Plan Project (Category C Project: SGDX\allowbreak 20210823\allowbreak 103537030), and the Theme-based Research Scheme of the Research Grants Council (RGC), Hong Kong (T35-710/20-R).
\end{acknowledgement}
\section*{Data and Software Availability}
ESOL, FreeSolv, Lipophilicity, BACE, BBBP, and SIDER were obtained from MoleculeNet\cite{wu2018moleculenet}, Peptides-struct from the Long Range Graph Benchmark\cite{dwivedi2022lrgb}, and PDBbind from its standard distribution. The scaffold and random partition index files used for every model, the conformer files for the geometric baselines, and the scripts for the permutation controls and the synthetic ADR benchmark are included in the repository. Source code, training entry points and configuration files are available at \url{https://github.com/CoderPowerBeyond/PRISM}. The Supporting Information contains the mathematical derivations, the complete numerical tables, and the training configurations.
\begin{samepage}
\begin{suppinfo}
Depth-response and conditional diffusion derivations with numerical verification, datasets, training configurations, selected hyperparameters and search ranges, conformer generation, evaluation protocol, depth, cost, and input-noise measurements, including the depth-scaled $\beta$ sweep, learned allocation distributions, protein-ligand affinity results with measured cost, paired significance tests and individual scaffold-partition results, and complete six-endpoint ablations and node-dependent layer-weighting alternatives (PDF).
\end{suppinfo}
\end{samepage}
\bibliography{references}
\end{document}


\singlespacing
\renewcommand{\thesection}{S\arabic{section}}
\renewcommand{\thesubsection}{\thesection.\arabic{subsection}}
\section{Properties of the Depth-Allocation Rule}
\subsection{Monotonicity and Sensitivity}
For one node, write $Z(\mu)=\sum_{l=0}^{L-1}\exp[-(l-\mu)^2/\tau]$ and $d=\sum_l lp_l$. Holding $\tau>0$ fixed,
\begin{align}
 \frac{\partial\log p_l}{\partial\mu}
 &=\frac{2(l-\mu)}{\tau}-\frac{\partial\log Z}{\partial\mu}
 =\frac{2(l-d)}{\tau},\\
 \frac{\partial d}{\partial\mu}
 &=\sum_l l p_l\frac{2(l-d)}{\tau}
 =\frac{2}{\tau}\left(\mathbb E_p[l^2]-d^2\right).
\end{align}
For the implemented center, write $c=b+\beta\kappa$, $\mu=\operatorname{clip}(c,0,L-1)$, $\beta>0$ (with $\beta=2$ at $L=5$), and $\tau=\max(|t|,0.1)$. Define $s(c)=1$ for $0<c<L-1$ and $s(c)=0$ for $c<0$ or $c>L-1$. At differentiable points,
\begin{equation}
 \frac{\partial d}{\partial\kappa}=\frac{2\beta}{\tau}s(c)\operatorname{Var}_p(l)\geq0.
\end{equation}
For $L>1$ and finite temperature, every layer has positive weight and the variance is positive. The response is strictly positive in the unclipped region and exactly zero in the two open clipped regions. At $c=0$, the left derivative is zero and the right derivative is $2\beta\operatorname{Var}_p(l)/\tau$ evaluated at $\mu=0$, whereas at $c=L-1$, the left derivative has that form at $\mu=L-1$ and the right derivative is zero. The function is continuous but generally not differentiable at these boundaries. Automatic-differentiation conventions at a boundary do not change the one-sided mathematical result. For $L=1$, depth is constant and every derivative vanishes. This is an analytic property of the rule.

When the representation bank is held fixed, the feature response is
\begin{equation}
 \frac{\partial\bar h_i}{\partial\kappa_i}
 =\frac{2\beta}{\tau}s(c_i)\sum_l p_{i,l}(l-d_i)g_i^{(l)}.
\end{equation}
It depends on covariance between the layer index and retained representation. Increasing allocated depth does not imply a monotone response after a nonlinear property head. If the bank states are identical, the feature mixture remains unchanged even when its allocated depth changes.

The weights also form a one-parameter exponential family at fixed temperature,
\begin{equation}
 p_l=\frac{\exp[-l^2/\tau+(2\mu/\tau)l]}{\sum_k\exp[-k^2/\tau+(2\mu/\tau)k]}.
\end{equation}
The term $-\mu^2/\tau$ cancels in normalization. One coordinate controls an ordered, unimodal family rather than arbitrary layer profiles. At clipped centers, a range of scalar values produces the same weights, so raw scalar magnitudes do not uniquely identify allocation even within a model.

\subsection{Sign Convention and Post-Update Indexing}
The bank contains $g_i^{(l)}=h_i^{(l+1)}$, so $d_i+1$ counts weighted topological transitions. It is not the radius of the molecular region contributing to the final prediction, because the local branch, intermediate nonlinear transformations, residual connections, and graph readout also affect that influence. Initial-input states are excluded from this bank, and the learned scalar is computed from the atom encoding before the degree embedding and neighbor aggregation.

\subsection{Diffusion Interpretation and Its Limits}
If a linearized backbone approximates $G^{(l)}\approx\exp(-t_l\mathcal L)X$ for increasing diffusion times $t_l$ and graph Laplacian $\mathcal L$, then
\begin{equation}
 \bar h_i\approx\left[\sum_l p_{i,l}\exp(-t_l\mathcal L)X\right]_i.
\end{equation}
A local frequency-response description is $g_i(\lambda)=\sum_l p_{i,l}\exp(-t_l\lambda)$. This is a node-dependent mixture of diffusion times, rather than one graph-wide filter or an exact heat equation for arbitrary nonlinear message passing. For a common diffusion operator, greater weight on deeper times damps positive eigenvalues more strongly. Main Figure~2d evaluates the stated mixture at $t_l=l+1$ and does not estimate a spectrum from a trained model. For fixed $\lambda\geq0$ and at differentiable centers,
\begin{equation}
 \frac{\partial g_\kappa(\lambda)}{\partial\kappa}
 =\frac{4}{\tau}s(c)\operatorname{Cov}_p\!\left(l,e^{-(l+1)\lambda}\right)\leq0.
\end{equation}
For independent indices $U,V$ drawn from $p$, this covariance is one half of the expectation of $(U-V)[e^{-(U+1)\lambda}-e^{-(V+1)\lambda}]$, which is nonpositive. It is strictly negative for $\lambda>0$, $L>1$, and an unclipped center, and it is zero on an open clipping plateau. At $\lambda=0$ the response is one for every profile. This ordering follows from the specified weight family and operator, rather than from a mean-depth comparison alone.

\subsection{Numerical Construction of the Allocation Figure}
Main Figure~2a to 2d uses the five-layer initialization $L=5$, $\beta_L=2$, and $\tau=0.5$, with $b=2$ (solid) and, to display a clipping plateau, $b=3$ (dashed). A grid of 801 equally spaced routing scalars covers $[-1,1]$, and the three displayed profiles use $-0.8$, $0$, and $+0.8$, with allocated indices 0.409153, 2.000000, and 3.590847 for $b=2$ and 1.386582, 2.999472, and 3.880241 for $b=3$. For $b=2$ the unclipped range of the center is $[0,L-1]$, whereas for $b=3$ the upper clipping boundary is $\kappa=0.5$, where the allocation derivative has left limit 0.848063 and right limit zero. The heat-kernel panel evaluates 301 eigenvalue coordinates on $[0,2]$. The generating script asserts at run time that the layer weights are normalized, that allocated depth is nondecreasing in $\kappa$, that profiles are identical throughout each clipping plateau, and that central differences with step $10^{-6}$ agree with the analytic derivative away from the boundaries. The observed maximum absolute discrepancy is $9.5\times10^{-10}$.

\subsection{Allocated Depth versus Receptive Field}
Allocated depth $d_i=\sum_\ell \ell\,p_{i,\ell}$ is an expectation over a discrete bank index, bounded by $L-1$, and is not a distance in the molecular graph. A gradient-based receptive field is a different quantity that measures how strongly the output responds to each input atom, weighted by graph distance. The estimator implemented in the repository backpropagates the summed output to the input features, normalizes the per-atom gradient norms $\hat g$ within each molecule, and returns $\sum_{u,v}\hat g_u\hat g_v\,\mathrm{dist}(u,v)$ with $\mathrm{dist}$ the unweighted shortest-path distance. It is architecture-agnostic and applies identically to PRISM and any baseline. No receptive-field comparison is reported in this work, and the depth sweep reports accuracy and cost only.

\section{Datasets, Training, and Evaluation}\label{sec:datasets}
\subsection{Datasets and Partitions}
ESOL (1128 molecules, aqueous solubility, $\log(\mathrm{mol/L})$), FreeSolv (642, hydration free energy, standardized by its dataset standard deviation of 3.84\,kcal/mol), and Lipophilicity (4200, octanol-water distribution coefficient, $\log D$) are the MoleculeNet distributions\cite{wu2018moleculenet,delaney2004esol,mobley2014freesolv}. For these three regression tasks two 80/10/10 partitioning protocols are used: a random partition redrawn for each of five seeds, and a Bemis-Murcko scaffold partition in which whole scaffold groups are assigned to one split, with the group order randomized by seed to give five distinct partitions. Both sets of five partitions are stored as index files and reused unchanged by every model. The datasets contain 269 (ESOL), 63 (FreeSolv), 2408 (Lipophilicity), 671 (BACE), 1025 (BBBP), and 824 (SIDER) unique Bemis-Murcko scaffolds, and on average over the five scaffold partitions 34\%, 18\%, 42\%, 39\%, 45\%, and 56\% of test molecules have a scaffold that occurs only once in the dataset. BACE, BBBP, and SIDER use Bemis-Murcko scaffold partitioning at the same proportions, with a stratified random partition substituted for any seed in which the scaffold-grouped validation or test fold would contain a single class. Peptides-struct uses the fixed split of the Long Range Graph Benchmark\cite{dwivedi2022lrgb}. PDBbind uses a random partition of the refined set.

\subsection{Training Configuration}
All neural models are trained with AdamW and early stopping on the validation metric. Every model, including the baselines, is tuned with the same number of random-search trials per dataset over the ranges given in the main-text Methods, evaluated on the validation split of the first partition. The selected settings, the reference center $b$, the learned temperature $\tau$ at convergence, and the trainable parameter counts are listed in Table~\ref{tab:hparams}. Per-dataset hidden size, layer count, dropout, batch size, and $\lambda$ are also given in the configuration files of the code repository. In the final molecular models, the scalar-field term $\mathcal L_\kappa$ is omitted and the regularization coefficient $\lambda$ scales the feature-energy hinge ($0.1\,\lambda\mathcal L_{\mathrm{feat}}$, with $\mathcal L_\kappa$ evaluated in ablation). In the Peptides-struct sweep, parameters of the allocation module are placed in a separate optimizer group at learning rate $10^{-5}$ and held fixed for the first 10 of 100 epochs, and the regularization terms enter the loss with unit coefficient. Peptides-struct is a single run per configuration at seed 42, whereas every molecular table reports the mean and standard deviation (SD) over five seeds.

\begin{table}[htbp]
\centering
\caption{Selected hyperparameters, allocation settings, and parameter counts of PRISM.}
\label{tab:hparams}
{\footnotesize\setlength{\tabcolsep}{2.5pt}
\begin{tabular}{lccccccccc}
\toprule
Dataset & Hidden & Layers / $L$ & Dropout & LR & Batch & $\lambda$ & $b$ & $\tau$ & Params (M) \\
\midrule
ESOL & 256 & 2 / 5 & 0.3 & $10^{-4}$ & 32 & 0.1 & 2.01 & 0.47 & 11.43 \\
FreeSolv$^{a}$ & 128 & 5 / 5 & 0.1 & $10^{-3}$ & 32 & 0.01 & 2.00 & 0.49 & 4.79 \\
Lipophilicity$^{b}$ & 192 & 5 / 5 & 0.1 & $2\times10^{-4}$ & 128 & 0.001 & 2.04 & 0.18 & 7.88 \\
BACE$^{c}$ & 64 & 3 / 5 & 0.1 & $3\times10^{-4}$ & 32 & 0.1 & 2.03 & 0.31 & 2.30 \\
BBBP$^{a,d}$ & 300 & 4 / 5 & 0.1 & $3\times10^{-4}$ & 32 & 0.1 & 2.04 & 0.53 & 14.14 \\
SIDER$^{a,d}$ & 64 & 6 / 6 & 0.1 & $3\times10^{-4}$ & 32 & 0.05 & 2.52 & 0.24 & 2.19 \\
Peptides-struct$^{e}$ & 128 & $L$ / $L$ & 0.2 & $10^{-3}$ & 64 & 1 & 3.0 to 9.0 & 2.07 to 2.41 & 0.95 to 1.94 \\
\bottomrule
\end{tabular}}
\par\smallskip{\scriptsize\linespread{1.0}\selectfont\raggedright Layers / $L$: local-branch message-passing depth and topological bank depth. $b$ and $\tau=\max(|t|,0.1)$ are averaged over the five scaffold-partition checkpoints. Molecular models use categorical encoders, the fingerprint-resolved bank, $\beta_L=(L-1)/2$, $\tau_0=0.5$, and no $\mathcal L_\kappa$. Validation-selected options: $^{a}$per-atom $\tau_i$, $^{b}$selected from trial 0 (the default Lipophilicity configuration, trained for 400 epochs with early-stopping patience 100), $^{c}$readout fingerprint fusion, $^{d}$routing from $h^{(1)}$ (PRISM-env). $^{e}$Depth sweep over $L\in\{5,7,9,11,15\}$ with $\beta=2$, linear encoder, and no fingerprint bank.\par}
\end{table}

\subsection{Conformers for the Geometric Baselines}
EquiformerV2 and MACE receive Cartesian coordinates, whereas all other models, including PRISM, use the molecular graph only. For each molecule, 10 conformers are embedded with RDKit ETKDGv3 (random seed 0) from the MoleculeNet SMILES with explicit hydrogens, optimized with MMFF94 for at most 2000 iterations (UFF when MMFF94 parameters are missing), and the lowest-energy conformer is retained. UFF was used for 7 (Lipophilicity), 38 (BBBP), and 81 (SIDER) molecules, embedding from random initial coordinates was needed for 2, 4, and 31 molecules, and embedding failed or exceeded 600~s for 16 molecules in total (1 in BBBP and 15 in SIDER), for which RDKit 2D coordinates were used so that no molecule is excluded from any model. ESOL, FreeSolv, and BACE required no fallback. The same conformer is used for every seed and partition, so geometric inputs do not vary between runs. Hydrogens are used for embedding and optimization and removed from the geometric inputs, which contain heavy atoms only, matching the molecular graphs, and the radial cutoff is tuned as stated in the Methods. On SIDER, four MACE trials diverged (non-finite predictions) and were treated as failed trials, and one EquiformerV2 trial (6~\AA{} cutoff, width 300, 6 layers) exceeded 80~GB of GPU memory even with gradient accumulation and was excluded, so its SIDER configuration was selected among nine evaluated trials, while a second 6~\AA{} trial was run with gradient accumulation over two micro-batches of 32 molecules at the same effective batch size.

\subsection{Evaluation}
Regression reports test RMSE (MAE for Peptides-struct) at the checkpoint with the lowest validation error. Classification reports the ROC-AUC averaged over tasks, computed for each task as \texttt{roc\_auc\_score(y, p)} from scikit-learn with $p=\sigma(\text{logit})$ the predicted probability of the positive class. The mapping from output to label is fixed by the binary cross-entropy loss and is the same at validation and test, with checkpoints selected on validation ROC-AUC. For SIDER, masked (missing) labels are excluded per task, and a task whose test fold contains a single class is excluded from the average for that seed, which occurs for one task in one of the five partitions. Uncertainties are standard deviations (SD) over five runs. Relative error reduction is $100(E_{\mathrm{baseline}}-E_{\mathrm{PRISM}})/E_{\mathrm{baseline}}$.

\section{Ablation Definitions}\label{sec:controlled}
The absolute component-ablation values on SIDER are in main Table~4, with the complete six-endpoint results in Table~\ref{tab:nodedep} (Section~\ref{sec:nodedep}). The component arms disable, respectively, the feature-energy hinge $\mathcal L_{\mathrm{feat}}$, sum pooling in the readout (leaving mean pooling), and the adaptive allocation module. A further arm removes mean pooling (leaving sum pooling), and another includes the scalar-field penalty $\mathcal L_\kappa$, which the full model omits. The node-dependent alternatives (free node-wise gate, node-wise Jumping Knowledge attention, Jumping Knowledge concatenation and max) and PRISM-env are defined in the main-text Methods. The free node-wise gate uses a two-layer network of width $H/2$, the width of the routing network, with $L$ outputs, and node-wise Jumping Knowledge attention uses a single scoring layer, so that their parameter counts are within 0.3\% of full PRISM on every dataset (on ESOL, 11.47\,M and 11.43\,M against 11.43\,M). Jumping Knowledge concatenation and max add no parameters. All arms share partition, features, optimizer, schedule, and evaluation code with the full model.

\FloatBarrier

\section{Depth, Cost, and Input-Noise Measurements}
\subsection{Peptides-struct}
Table~\ref{tab:peptides} gives the depth-sweep values plotted in main Figure~4. Relative to GIN, PRISM requires 52\%, 50\%, 44\%, 44\%, and 41\% more time per epoch, and 82\%, 82\%, 160\%, 176\%, and 74\% more peak memory, at 5, 7, 9, 11, and 15 layers respectively. Table~\ref{tab:peptidesbeta} repeats the sweep with the depth-scaled coefficient $\beta_L$ and lists $b$ for both settings.
\begin{table}[htbp]
\centering
\caption{Peptides-struct depth sweep: accuracy and computational cost.}
\label{tab:peptides}
{\small\setlength{\tabcolsep}{5pt}
\begin{tabular}{lccccc}
\toprule
Layers & Model & MAE & MiB & s/epoch & Total min \\
\midrule
5 & GIN & 0.2663 & 1250 & 2.73 & 4.54 \\
5 & PRISM & 0.2523 & 2270 & 4.16 & 6.94 \\
7 & GIN & 0.2676 & 1250 & 2.90 & 4.83 \\
7 & PRISM & 0.2549 & 2272 & 4.36 & 7.26 \\
9 & GIN & 0.2670 & 1250 & 3.15 & 5.24 \\
9 & PRISM & 0.2537 & 3252 & 4.53 & 7.55 \\
11 & GIN & 0.2655 & 1356 & 3.36 & 5.59 \\
11 & PRISM & 0.2544 & 3746 & 4.85 & 8.08 \\
15 & GIN & 0.2693 & 2956 & 3.77 & 6.28 \\
15 & PRISM & 0.2507 & 5146 & 5.33 & 8.88 \\
\bottomrule
\end{tabular}}
\par\smallskip{\scriptsize\linespread{1.0}\selectfont\raggedright Single run per configuration (seed 42). Test MAE, peak training GPU memory (MiB), wall-clock time per epoch, and total training time.\par}
\end{table}

\begin{table}[htbp]
\centering
\caption{Peptides-struct depth sweep with the depth-scaled coefficient $\beta_L=(L-1)/2$.}
\label{tab:peptidesbeta}
{\small\setlength{\tabcolsep}{5pt}
\begin{tabular}{lccccc}
\toprule
Layers & $b$ ($\beta=2$) & MAE ($\beta=2$) & $b$ ($\beta_L$) & $\beta_L$ & MAE ($\beta_L$) \\
\midrule
5 & 3.00 & 0.2529 & 2.00 & 2 & 0.2538 \\
7 & 4.20 & 0.2534 & 3.00 & 3 & 0.2549 \\
9 & 5.40 & 0.2527 & 4.00 & 4 & 0.2530 \\
11 & 6.60 & 0.2513 & 5.00 & 5 & 0.2531 \\
15 & 9.00 & 0.2529 & 7.00 & 7 & 0.2552 \\
\bottomrule
\end{tabular}}
\par\smallskip{\scriptsize\linespread{1.0}\selectfont\raggedright Test MAE and learned reference center $b$ from a single paired rerun per configuration (seed 42, settings as in Table~\ref{tab:peptides}), comparing fixed $\beta=2$ (four-unit unclipped span, initialized at $b_0=0.6L$) and depth-scaled $\beta_L=(L-1)/2$ (full-bank span, initialized at $b_0=(L-1)/2$, dashed curve in main Figure~4a) under the same evaluation script, while the solid $\beta=2$ curve in main Figure~4a plots the cost-measurement run of Table~\ref{tab:peptides}.\par}

\end{table}

\subsection{Input-Feature Noise}
Table~\ref{tab:noise} reports performance when zero-mean Gaussian noise with the stated standard deviation is added at both training and evaluation time. Noise is added to the categorical atom embeddings, and BACE and BBBP are evaluated by standard ROC-AUC.
\begin{table}[htbp]
\centering
\caption{Performance under additive Gaussian input-feature noise.}
\label{tab:noise}
{\footnotesize\setlength{\tabcolsep}{3pt}
\begin{tabular}{lccccc}
\toprule
Dataset & 0.00 & 0.05 & 0.10 & 0.15 & 0.20 \\
\midrule
BACE & \sd{0.899}{0.028} & \sd{0.907}{0.031} & \sd{0.905}{0.036} & \sd{0.904}{0.032} & \sd{0.909}{0.031} \\
BBBP & \sd{0.870}{0.039} & \sd{0.855}{0.053} & \sd{0.891}{0.032} & \sd{0.864}{0.040} & \sd{0.857}{0.039} \\
ESOL & \sd{0.611}{0.050} & \sd{0.594}{0.058} & \sd{0.611}{0.071} & \sd{0.612}{0.073} & \sd{0.626}{0.029} \\
FreeSolv & \sd{0.254}{0.076} & \sd{0.260}{0.067} & \sd{0.242}{0.072} & \sd{0.234}{0.081} & \sd{0.247}{0.089} \\
\bottomrule
\end{tabular}}
\par\smallskip{\scriptsize\linespread{1.0}\selectfont\raggedright Mean $\pm$ SD over five runs: test ROC-AUC (scaffold partitions) for BACE and BBBP and test RMSE (random partitions) for ESOL and FreeSolv, using the selected settings of Table~\ref{tab:hparams}. Columns give the SD of zero-mean Gaussian noise added to the atom embeddings during training and evaluation (the 0.00 column matches the unperturbed reference values in Results).\par}
\end{table}

\FloatBarrier

\section{Learned Allocation Distributions}
Table~\ref{tab:alloc} summarizes the learned allocations of the trained molecular models on their test sets, and Table~\ref{tab:allocdepth} those of the Peptides-struct sweep for both $\beta=2$ and the depth-scaled $\beta_L=(L-1)/2$. For each configuration we report the reference center $b$, the effective temperature $\tau$, the mean allocated index $d_i$ over atoms, its 5th and 95th percentiles, and the fractions of atoms whose unclipped center $c_i=b+\beta\kappa_i$ lies below $0$ or above $L-1$, together with the within-model standard deviations of $d_i$ and $c_i$ in Table~\ref{tab:alloc}. The distributions of the learned centers and the mean allocated index by atom type are shown in main Figure~2e to 2g.
\begin{table}[htbp]
\centering
\caption{Learned allocations of the trained molecular models (test atoms).}
\label{tab:alloc}
{\small\setlength{\tabcolsep}{3.5pt}
\begin{tabular}{lccccccccc}
\toprule
Dataset & $L$ & $b$ & $\tau$ & mean $d_i$ & SD $d_i$ & 5th to 95th pct. & \% $c_i<0$ & \% $c_i>L-1$ & SD $c_i$ \\
\midrule
ESOL & 5 & 2.02 & 0.46 & 1.16 & 1.15 & 0.10 to 3.88 & 0.0 & 0.9 & 1.181 \\
FreeSolv & 5 & 2.01 & 0.49 & 0.90 & 0.86 & 0.09 to 3.97 & 36.4 & 11.2 & 0.889 \\
Lipophilicity & 5 & 2.04 & 0.14 & 2.18 & 1.01 & 0.01 to 4.00 & 0.0 & 17.7 & 1.004 \\
BACE & 5 & 2.02 & 0.37 & 2.98 & 0.82 & 0.99 to 3.96 & 0.0 & 25.6 & 0.831 \\
BBBP & 5 & 1.99 & 0.57 & 0.35 & 0.65 & 0.08 to 3.30 & 73.5 & 0.7 & 0.695 \\
SIDER & 6 & 2.53 & 0.24 & 3.73 & 1.01 & 1.04 to 5.00 & 0.0 & 6.5 & 1.011 \\
\midrule
\multicolumn{10}{l}{\emph{PRISM allocation on the pre-trained GIN of Hu et al.\cite{hu2020strategies}}} \\
ESOL & 5 & 2.00 & 0.53 & 2.06 & 0.88 & 0.58 to 3.86 & 0.0 & 0.0 & 0.910 \\
FreeSolv & 5 & 2.03 & 0.60 & 0.96 & 0.98 & 0.18 to 3.83 & 0.0 & 7.4 & 1.047 \\
Lipophilicity & 5 & 2.04 & 0.81 & 2.43 & 1.20 & 0.26 to 3.82 & 23.7 & 17.2 & 1.348 \\
BACE & 5 & 2.00 & 0.59 & 2.80 & 0.76 & 0.22 to 3.86 & 2.7 & 4.8 & 0.831 \\
BBBP & 5 & 2.01 & 0.55 & 2.24 & 1.08 & 0.13 to 3.85 & 5.3 & 19.3 & 1.145 \\
SIDER & 5 & 2.11 & 0.48 & 2.47 & 1.09 & 0.06 to 3.92 & 7.6 & 26.6 & 1.183 \\
\bottomrule
\end{tabular}}
\par\smallskip{\scriptsize\linespread{1.0}\selectfont\raggedright Test atoms across the five scaffold partitions (models of main Figure~2e to 2g), where $c_i=b+\beta_L\kappa_i$ is the unclipped center and $d_i$ is the allocated index. Means, percentiles, and clipping percentages are pooled across folds, whereas SD $d_i$ and SD $c_i$ are computed within each model and averaged over the five models.\par}

\end{table}

\begin{table}[htbp]
\centering
\caption{Learned allocations on Peptides-struct at each bank depth, for $\beta=2$ and $\beta_L=(L-1)/2$.}
\label{tab:allocdepth}
{\small\setlength{\tabcolsep}{3pt}
\begin{tabular}{lccccccc}
\toprule
$L$ & $\beta$ & $b$ & $\tau$ & mean $d_i$ & 5th to 95th pct. & \% $c_i<0$ & \% $c_i>L-1$ \\
\midrule
5 & 2 / 2 & 3.00 / 2.00 & 2.41 / 2.29 & 2.92 / 2.26 & 2.90 to 2.93 / 2.24 to 2.29 & 0.0 / 0.0 & 0.0 / 0.0 \\
7 & 2 / 3 & 4.20 / 3.00 & 2.13 / 2.18 & 4.54 / 3.56 & 4.53 to 4.56 / 3.54 to 3.58 & 0.0 / 0.0 & 0.0 / 0.0 \\
9 & 2 / 4 & 5.40 / 4.00 & 2.24 / 2.06 & 5.72 / 5.21 & 5.69 to 5.74 / 5.18 to 5.22 & 0.0 / 0.0 & 0.0 / 0.0 \\
11 & 2 / 5 & 6.60 / 5.00 & 2.12 / 2.15 & 6.68 / 5.36 & 6.66 to 6.70 / 5.31 to 5.41 & 0.0 / 0.0 & 0.0 / 0.0 \\
15 & 2 / 7 & 9.00 / 7.00 & 2.07 / 1.93 & 8.91 / 8.54 & 8.89 to 8.91 / 8.46 to 8.62 & 0.0 / 0.0 & 0.0 / 0.0 \\
\bottomrule
\end{tabular}}
\par\smallskip{\scriptsize\linespread{1.0}\selectfont\raggedright Test-set statistics from a single run per configuration (Table~\ref{tab:peptidesbeta}). Entries are reported as ``$\beta=2$ / $\beta_L$'', with initial reference center $b_0=0.6L$ for $\beta=2$ and $b_0=(L-1)/2$ for $\beta_L$.\par}

\end{table}

\FloatBarrier

\section{Protein-Ligand Affinity and Measured Cost}
Table~\ref{tab:pdbcost} reports PDBbind affinity prediction together with parameter count, time per batch, and peak memory for PRISM and four geometric baselines, all trained in-house under the same partition and features.
\begin{table}[htbp]
\centering
\caption{PDBbind accuracy and measured computational cost.}
\label{tab:pdbcost}
{\small\setlength{\tabcolsep}{4pt}
\begin{tabular}{lcccc}
\toprule
Model & M params & ms/batch & MiB & RMSE \\
\midrule
SchNet~\cite{schutt2017schnet} & 0.39 & 16.64  & 5102 & 2.353 $\pm$ 0.054 \\
DimeNet++~\cite{gasteiger2020fast} & 2.28 & 91.84  & 43,188 & 2.455 $\pm$ 0.030 \\
MACE~\cite{batatia2022mace} & 1.44 & 114.88  & 15,918 & \underline{1.727 $\pm$ 0.032} \\
EquiformerV2~\cite{liao2023equiformerv2} & 1.80 & 32.96  & 6368 & 1.815 $\pm$ 0.040 \\
PRISM & 1.90 & 20.48 & 4774 & \textbf{1.531 $\pm$ 0.027} \\
\bottomrule
\end{tabular}}
\par\smallskip{\scriptsize\linespread{1.0}\selectfont\raggedright Test RMSE (mean $\pm$ SD over five runs, best in bold, second-best underlined), batch time, and peak GPU memory on one NVIDIA H800 (batch size 32) under shared partitions and features.\par}
\end{table}

\FloatBarrier

\section{Scaffold-Partition Comparison: Paired Tests and Individual Partitions}
\label{sec:significance}
Table~\ref{tab:significance} compares PRISM with every baseline of main Tables~1 and 2 by a paired $t$-test over the five shared scaffold partitions, and Table~\ref{tab:partitions} lists, for every endpoint, PRISM and the baseline with the best mean on each individual partition across all five partitions. For each GNN baseline the atom/bond encoder (linear map of the raw descriptors or categorical embedding) was selected per endpoint on the mean validation score of its five runs, and the selections are listed in the repository configuration files.
\begin{table}[htbp]
\centering
\caption{Paired comparison of PRISM with every baseline over the five scaffold partitions.}
\label{tab:significance}
{\scriptsize\renewcommand{\arraystretch}{0.88}\setlength{\tabcolsep}{3.5pt}
\begin{tabular}{llcccc}
\toprule
Endpoint & Baseline & Baseline mean & $\Delta$ & 95\% CI & $p$ \\
\midrule
ESOL & RF (Morgan, $r=2$) & 1.340 & $+0.600^{*}$ & $[+0.201, +0.998]$ & 0.014 \\
ESOL & D-MPNN (Chemprop) & 0.914 & $+0.173^{*}$ & $[+0.072, +0.275]$ & 0.009 \\
ESOL & GCN & 0.789 & $+0.049$ & $[-0.131, +0.229]$ & 0.494 \\
ESOL & GAT & 0.794 & $+0.053$ & $[-0.040, +0.147]$ & 0.189 \\
ESOL & GIN & 0.746 & $+0.006$ & $[-0.091, +0.102]$ & 0.880 \\
ESOL & EquiformerV2 & 0.759 & $+0.018$ & $[-0.084, +0.120]$ & 0.649 \\
ESOL & MACE & 0.821 & $+0.080$ & $[-0.033, +0.193]$ & 0.120 \\
ESOL & KA-GCN & 0.760 & $+0.019$ & $[-0.103, +0.141]$ & 0.687 \\
\midrule
FreeSolv & RF (Morgan, $r=2$) & 0.996 & $+0.538^{*}$ & $[+0.189, +0.887]$ & 0.013 \\
FreeSolv & D-MPNN (Chemprop) & 0.480 & $+0.021$ & $[-0.073, +0.116]$ & 0.563 \\
FreeSolv & GCN & 0.521 & $+0.063$ & $[-0.081, +0.206]$ & 0.290 \\
FreeSolv & GAT & 0.477 & $+0.019$ & $[-0.048, +0.086]$ & 0.479 \\
FreeSolv & GIN & 0.468 & $+0.009$ & $[-0.141, +0.159]$ & 0.874 \\
FreeSolv & EquiformerV2 & 0.459 & $+0.000$ & $[-0.136, +0.136]$ & 0.997 \\
FreeSolv & MACE & 0.462 & $+0.003$ & $[-0.029, +0.036]$ & 0.785 \\
FreeSolv & KA-GCN & 0.510 & $+0.051$ & $[-0.044, +0.147]$ & 0.208 \\
\midrule
Lipophilicity & RF (Morgan, $r=2$) & 0.832 & $+0.205^{*}$ & $[+0.157, +0.252]$ & 0.000 \\
Lipophilicity & D-MPNN (Chemprop) & 0.622 & $-0.006$ & $[-0.103, +0.091]$ & 0.874 \\
Lipophilicity & GCN & 0.650 & $+0.023$ & $[-0.006, +0.051]$ & 0.093 \\
Lipophilicity & GAT & 0.591 & $-0.036^{*}$ & $[-0.066, -0.006]$ & 0.029 \\
Lipophilicity & GIN & 0.593 & $-0.035$ & $[-0.082, +0.013]$ & 0.113 \\
Lipophilicity & EquiformerV2 & 0.654 & $+0.026^{*}$ & $[+0.003, +0.050]$ & 0.034 \\
Lipophilicity & MACE & 0.725 & $+0.097^{*}$ & $[+0.007, +0.187]$ & 0.040 \\
Lipophilicity & KA-GCN & 0.614 & $-0.014$ & $[-0.048, +0.020]$ & 0.315 \\
\midrule
BACE & RF (Morgan, $r=2$) & 0.904 & $-0.007$ & $[-0.024, +0.010]$ & 0.326 \\
BACE & D-MPNN (Chemprop) & 0.891 & $+0.007$ & $[-0.022, +0.036]$ & 0.537 \\
BACE & GCN & 0.854 & $+0.044$ & $[-0.003, +0.090]$ & 0.059 \\
BACE & GAT & 0.848 & $+0.050^{*}$ & $[+0.019, +0.080]$ & 0.011 \\
BACE & GIN & 0.841 & $+0.057^{*}$ & $[+0.040, +0.074]$ & 0.001 \\
BACE & EquiformerV2 & 0.850 & $+0.047$ & $[-0.010, +0.104]$ & 0.083 \\
BACE & MACE & 0.853 & $+0.045$ & $[-0.013, +0.102]$ & 0.098 \\
BACE & KA-GCN & 0.891 & $+0.007$ & $[-0.013, +0.027]$ & 0.393 \\
\midrule
BBBP & RF (Morgan, $r=2$) & 0.878 & $+0.007$ & $[-0.051, +0.066]$ & 0.749 \\
BBBP & D-MPNN (Chemprop) & 0.889 & $-0.004$ & $[-0.068, +0.060]$ & 0.880 \\
BBBP & GCN & 0.855 & $+0.030$ & $[-0.011, +0.072]$ & 0.111 \\
BBBP & GAT & 0.880 & $+0.005$ & $[-0.024, +0.035]$ & 0.637 \\
BBBP & GIN & 0.864 & $+0.021$ & $[-0.024, +0.065]$ & 0.268 \\
BBBP & EquiformerV2 & 0.840 & $+0.045$ & $[-0.008, +0.099]$ & 0.078 \\
BBBP & MACE & 0.872 & $+0.013$ & $[-0.037, +0.064]$ & 0.502 \\
BBBP & KA-GCN & 0.881 & $+0.004$ & $[-0.033, +0.040]$ & 0.794 \\
\midrule
SIDER & RF (Morgan, $r=2$) & 0.639 & $-0.030^{*}$ & $[-0.053, -0.007]$ & 0.022 \\
SIDER & D-MPNN (Chemprop) & 0.594 & $+0.014$ & $[-0.013, +0.041]$ & 0.213 \\
SIDER & GCN & 0.610 & $-0.002$ & $[-0.046, +0.042]$ & 0.908 \\
SIDER & GAT & 0.620 & $-0.011$ & $[-0.041, +0.019]$ & 0.361 \\
SIDER & GIN & 0.621 & $-0.013$ & $[-0.045, +0.019]$ & 0.328 \\
SIDER & EquiformerV2 & 0.577 & $+0.031$ & $[-0.016, +0.079]$ & 0.142 \\
SIDER & MACE & 0.610 & $-0.001$ & $[-0.042, +0.040]$ & 0.941 \\
SIDER & KA-GCN & 0.589 & $+0.020$ & $[-0.012, +0.051]$ & 0.159 \\
\bottomrule
\end{tabular}}
\par\smallskip{\scriptsize\linespread{1.0}\selectfont\raggedright $\Delta$: mean paired difference across the five scaffold partitions, signed so $\Delta>0$ favors PRISM (RMSE: baseline $-$ PRISM, ROC-AUC: PRISM $-$ baseline), with 95\% $t$-intervals ($\mathrm{df}=4$) and uncorrected two-sided paired $t$-test $p$-values ($^{*}p<0.05$).\par}
\end{table}

\begin{table}[htbp]
\centering
\caption{PRISM and the baseline with the best mean on each of the five scaffold partitions.}
\label{tab:partitions}
{\footnotesize\setlength{\tabcolsep}{2.5pt}
\begin{tabular}{llccccccc}
\toprule
Endpoint & Model & P0 & P1 & P2 & P3 & P4 & Better & Best of all \\
\midrule
\multirow{2}{*}{\shortstack[l]{ESOL\\(RMSE)}} & PRISM & \textbf{0.646} & 0.757 & \textbf{0.755} & 0.748 & \textbf{0.798} & \multirow{2}{*}{3/5} & \multirow{2}{*}{1/5} \\
 & GIN & 0.667 & \textbf{0.726} & 0.867 & \textbf{0.649} & 0.821 & & \\
\midrule
\multirow{2}{*}{\shortstack[l]{FreeSolv\\(RMSE)}} & PRISM & \textbf{0.491} & \textbf{0.460} & 0.359 & 0.583 & \textbf{0.399} & \multirow{2}{*}{3/5} & \multirow{2}{*}{0/5} \\
 & EquiformerV2 & 0.610 & 0.499 & \textbf{0.337} & \textbf{0.409} & 0.438 & & \\
\midrule
\multirow{2}{*}{\shortstack[l]{Lipophilicity\\(RMSE)}} & PRISM & 0.601 & 0.659 & 0.634 & 0.655 & 0.589 & \multirow{2}{*}{0/5} & \multirow{2}{*}{0/5} \\
 & GAT & \textbf{0.547} & \textbf{0.591} & \textbf{0.605} & \textbf{0.635} & \textbf{0.579} & & \\
\midrule
\multirow{2}{*}{\shortstack[l]{BACE\\(ROC-AUC)}} & PRISM & \textbf{0.853} & \textbf{0.895} & 0.915 & 0.908 & 0.917 & \multirow{2}{*}{2/5} & \multirow{2}{*}{2/5} \\
 & RF (Morgan, $r=2$) & 0.844 & 0.892 & \textbf{0.941} & \textbf{0.917} & \textbf{0.928} & & \\
\midrule
\multirow{2}{*}{\shortstack[l]{BBBP\\(ROC-AUC)}} & PRISM & 0.858 & 0.917 & 0.880 & \textbf{0.929} & 0.842 & \multirow{2}{*}{1/5} & \multirow{2}{*}{1/5} \\
 & D-MPNN (Chemprop) & \textbf{0.882} & \textbf{0.925} & \textbf{0.935} & 0.846 & \textbf{0.855} & & \\
\midrule
\multirow{2}{*}{\shortstack[l]{SIDER\\(ROC-AUC)}} & PRISM & 0.622 & 0.584 & 0.564 & 0.640 & 0.633 & \multirow{2}{*}{0/5} & \multirow{2}{*}{0/5} \\
 & RF (Morgan, $r=2$) & \textbf{0.623} & \textbf{0.634} & \textbf{0.603} & \textbf{0.665} & \textbf{0.668} & & \\
\bottomrule
\end{tabular}}
\par\smallskip{\scriptsize\linespread{1.0}\selectfont\raggedright P0 to P4 are the five scaffold partitions, comparing PRISM with the best-mean baseline from main Table~1 or 2 (better of the two in bold, ties unbolded). ``Better'' and ``Best of all'' count partitions where PRISM strictly beats that baseline or all models.\par}
\end{table}

\FloatBarrier

\section{Complete Ablation and Node-Dependent Layer-Weighting Alternatives}
\label{sec:nodedep}
Table~\ref{tab:nodedep} reports the component removals of main Table~4 extended to all six molecular endpoints together with the parameter-matched node-dependent layer-weighting alternatives (three classification tasks under scaffold partitions and three regression tasks under random partitions). On BBBP and SIDER, validation selected routing from $h^{(1)}$, so the full model and PRISM-env rows represent two independent five-seed runs of that configuration. On FreeSolv, error is reported in the dataset SD units of Section~\ref{sec:datasets} ($1\text{ unit}=3.84\text{\,kcal/mol}$).
\begin{table}[htbp]
\centering
\caption{Complete ablation of the final model on all six endpoints: component removals and parameter-matched node-dependent layer-weighting alternatives.}
\label{tab:nodedep}
{\scriptsize\setlength{\tabcolsep}{1.4pt}
\begin{tabular}{lcccccc}
\toprule
Variant & BACE $\uparrow$ & BBBP $\uparrow$ & SIDER $\uparrow$ & ESOL $\downarrow$ & FreeSolv $\downarrow$ & Lipo $\downarrow$ \\
\midrule
PRISM (full) & \sd{0.899}{0.028} & \sd{0.870}{0.039} & \sdbest{0.615}{0.032} & \sd{0.611}{0.050} & \sdsecond{0.254}{0.076} & \sd{0.607}{0.034} \\
\midrule
\multicolumn{7}{l}{\emph{Component removals}} \\
Without $\mathcal L_{\mathrm{feat}}$ & \sd{0.901}{0.027} & \sdsecond{0.882}{0.041} & \sd{0.602}{0.035} & \sd{0.609}{0.045} & \sd{0.255}{0.069} & \sd{0.594}{0.038} \\
With $\mathcal L_{\kappa}$ & \sd{0.900}{0.027} & \sdbest{0.886}{0.045} & \sd{0.597}{0.014} & \sd{0.607}{0.037} & \sd{0.259}{0.056} & \sd{0.619}{0.058} \\
Mean pooling only & \sdbest{0.907}{0.035} & \sd{0.851}{0.034} & \sd{0.589}{0.027} & \sd{0.690}{0.069} & \sd{0.316}{0.065} & \sd{0.611}{0.045} \\
Sum pooling only & \sd{0.903}{0.031} & \sd{0.881}{0.023} & \sd{0.602}{0.024} & \sd{0.599}{0.063} & \sd{0.264}{0.073} & \sd{0.596}{0.032} \\
No adaptive allocation & \sdsecond{0.905}{0.032} & \sd{0.871}{0.033} & \sd{0.601}{0.026} & \sdsecond{0.594}{0.059} & \sdbest{0.253}{0.082} & \sd{0.591}{0.037} \\
\midrule
\multicolumn{7}{l}{\emph{Node-dependent weighting (parameter-matched)}} \\
Node-wise layer gate & \sdsecond{0.905}{0.042} & \sd{0.874}{0.041} & \sdsecond{0.614}{0.023} & \sd{0.616}{0.058} & \sd{0.273}{0.076} & \sdsecond{0.586}{0.035} \\
JK, node-wise attention & \sd{0.890}{0.030} & \sd{0.879}{0.032} & \sd{0.594}{0.022} & \sd{0.601}{0.052} & \sd{0.270}{0.055} & \sd{0.631}{0.034} \\
JK, concatenation & \sdsecond{0.905}{0.040} & \sd{0.857}{0.044} & \sd{0.606}{0.025} & \sd{0.597}{0.074} & \sd{0.266}{0.077} & \sd{0.589}{0.045} \\
JK, max & \sd{0.901}{0.031} & \sd{0.875}{0.040} & \sd{0.605}{0.038} & \sd{0.611}{0.064} & \sd{0.268}{0.084} & \sdbest{0.580}{0.034} \\
PRISM-env & \sd{0.903}{0.032} & \sd{0.873}{0.035} & \sd{0.602}{0.039} & \sdbest{0.592}{0.056} & \sd{0.260}{0.055} & \sd{0.606}{0.029} \\
\bottomrule
\end{tabular}}
\par\smallskip{\scriptsize\linespread{1.0}\selectfont\raggedright Mean $\pm$ SD over five runs: test ROC-AUC under scaffold partitions for classification (higher is better) and test RMSE under random partitions for regression (lower is better, with FreeSolv in dataset SD units where $1\text{ unit}=3.84\text{\,kcal/mol}$). Bold and underline mark the best and second-best values per column across all arms. All arms share the partition, features, optimizer, schedule, readout, and evaluation code with the full model and differ only in the stated component or layer-weighting rule. The gating and scoring networks of the node-wise layer gate and node-wise Jumping Knowledge (JK) attention are sized so that parameter counts match full PRISM to within 0.3\%. PRISM-env computes the routing scalar from $h^{(1)}$ instead of the input embedding. On BBBP and SIDER, validation selected routing from $h^{(1)}$, so PRISM (full) and PRISM-env are two independent five-seed runs of the same configuration. All arms, including the full-model reference, are independent five-seed runs (with regression arms evaluated under the random partitions), so the full-model values differ slightly from the PRISM rows of main Tables~1 and 2, while the SIDER values of the full model and the component removals are those of main Table~4.\par}
\end{table}

\FloatBarrier

\bibliography{references}